\documentclass[runningheads]{llncs}

\usepackage{eccv}
\usepackage{eccvabbrv}

\usepackage{graphicx}
\usepackage{booktabs}
\usepackage{fontawesome5}

\usepackage[accsupp]{axessibility} 
\usepackage{tabularray}
\UseTblrLibrary{booktabs}
\usepackage[table,dvipsnames]{xcolor} 
\usepackage{xparse}
\usepackage{tikz}
\usetikzlibrary{positioning,spy,calc} 
\usepackage{wrapfig}
\colorlet{HeaderGray}{white!94!black}
\colorlet{LightGray}{White!90!Periwinkle}
\colorlet{BandGreen}{green!12}
\colorlet{OursGreen}{green!8}
\colorlet{OursSoft}{pink!15}
\definecolor{tealnum}{HTML}{008A8C}
\colorlet{LightBeige}{green!5}
\definecolor{Oatmeal}{HTML}{F2EFE9}
\colorlet{PeachBeige}{blue!5}
\definecolor{deepblue}{RGB}{23,53,122}
\definecolor{deepred}{RGB}{139,26,26}
\definecolor{colar}{RGB}{249,222,213}
\usepackage{pifont}
\definecolor{CheckTeal}{HTML}{3D8B86}
\definecolor{CrossCoral}{HTML}{B85C5A}
\newcommand{\cmark}{\textcolor{CheckTeal}{\ding{51}}}
\newcommand{\xmark}{\textcolor{CrossCoral}{\ding{55}}}

\usepackage[breaklinks,colorlinks,linkcolor=deepred,urlcolor=deepblue,citecolor=tealnum]{hyperref} 
\usepackage{orcidlink}

\begin{document}

\title{Recurrent Sinusoidal INRs for Efficient High-Fidelity Representation}

\titlerunning{Harmonic-line Spectrum INR}

\author{Hyunmin Cho\inst{1}\orcidlink{0009-0003-6199-9751} \and
Jaejun Yoo\inst{2}\orcidlink{0000-0001-5252-9668} \and
Kyong Hwan Jin\inst{1}\thanks{Correspondence to: Kyong Hwan Jin \textless{kyong\_jin@korea.ac.kr}\textgreater.}\orcidlink{0000-0001-7885-4792}}

\authorrunning{H.~Cho et al.}

\institute{Department of Electrical Engineering, Korea University, Seoul, South Korea\\
\email{\{hyun\_cho,kyong\_jin\}@korea.ac.kr}\and
Graduate School of Artificial Intelligence, UNIST, Ulsan, South Korea\\
\email{jaejun.yoo@unist.ac.kr }\\
\faGithub ~Project Page: \href{https://hyeon-cho.github.io/Harmonic-line-Spectrum/} {\texttt{hyeon-cho.github.io/Harmonic-line-Spectrum/}}
}

\maketitle

\begin{abstract}
We study sinusoidal recurrence as an iterative mechanism for harmonic spectral enrichment in implicit neural representations (INRs). Our analysis reveals that sinusoidal activations induce a harmonic line spectrum, providing a spectral account of how recurrent unrolling enriches the effective spectral support. We realize this principle with a shared sinusoidal block that iteratively refines the latent representation. We empirically validate the resulting spectral behavior against feed-forward INRs, non-sinusoidal recurrent variants, and equilibrium-style sinusoidal models. Complementing this analysis, we evaluate the proposed architecture across image and 3D representation tasks. On RGB image benchmarks, our method achieves higher fidelity than feed-forward baselines with fewer parameters and fewer optimization steps, and it further transfers favorably to super-resolution, NeRF, and SDF tasks.
\keywords{Recurrent network \and Implicit signal representation}
\end{abstract}

\section{Introduction}
\label{sec:intro}
We study implicit neural representations (INRs)~\cite{Park_2019_CVPR,10.1007/978-3-030-58452-8_24,NEURIPS2020_53c04118,Liu_2024_CVPR,Saragadam_2023_CVPR}, which parameterize a signal as a neural function $f$, typically a coordinate-based MLP,{
\setlength{\abovedisplayskip}{3pt}
\setlength{\belowdisplayskip}{3pt}
\[
    f:\mathbb{R}^d \rightarrow \mathbb{R}^c. \nonumber
\]}By treating the input as spatial or spatio-temporal coordinates, INRs represent continuous signals and can interpolate values at unseen locations. This property has made them a standard tool in applications such as 3D reconstruction and novel view synthesis~\cite{Park_2019_CVPR,10.1007/978-3-030-58452-8_24,Mescheder_2019_CVPR}.

A central challenge of coordinate-based MLPs is \emph{spectral bias}: neural networks tend to fit low-frequency structure more readily than high-frequency structure, making faithful recovery of fine detail difficult~\cite{pmlr-v97-rahaman19a,NEURIPS2020_55053683}. A large body of work has addressed this issue through activation design, input encodings, multiresolution parameterizations, and optimization strategies~\cite{NEURIPS2020_53c04118,10.1145/3528223.3530127,Liu_2024_CVPR,Han_2025_CVPR,mcginnis2025optimizing}. While effective, many of these approaches improve fidelity by increasing model size, auxiliary parameters, or training complexity.

To reason about where high-frequency information is gained or lost, it is useful to decompose a typical INR into three stages. For a target signal $\mathbf{s}$ evaluated at coordinates $\mathbf{x}$, we write{
\setlength{\abovedisplayskip}{5pt}
\setlength{\belowdisplayskip}{5pt}
\begin{equation}
    \mathbf{s}
    =
    \mathcal{D}\!\left(\mathcal{F}\!\left(\mathcal{E}(\mathbf{x})\right)\right),
    \label{eq:inr_decomp}
\end{equation}}where $\mathcal{E}$ lifts low-dimensional coordinates into a higher-dimensional representation~\cite{NEURIPS2020_55053683,10.1145/3528223.3530127}, $\mathcal{F}$ transforms this representation into a latent feature, and $\mathcal{D}$ maps the latent feature to the output space. Under this view, high-fidelity reconstruction depends not only on the coordinate features introduced by $\mathcal{E}$, but also on how $\mathcal{F}$ composes and refines them into representations capable of resolving fine-scale structure.

A straightforward way to improve the ability of $\mathcal{F}$ to resolve fine-scale structure is to increase its representational capacity through deeper or wider latent transformations. While effective, this approach introduces additional independently parameterized layers and increased computation. This leaves open a complementary question: \emph{can $\mathcal{F}$ be made more effective at resolving fine-scale structure without increasing independently parameterized depth?}

Recurrence provides a natural mechanism for such refinement. Unrolling a recurrent update for $T$ steps increases effective depth while reusing the same parameters, offering a parameter-efficient alternative to independently parameterized depth~\cite{ELMAN1990179,NEURIPS2019_01386bd6,schwarzschild2022the}. Related INR formulations have also explored iterative latent computation through equilibrium networks, whose fixed-point formulation enables constant-memory backpropagation with respect to effective depth~\cite{
NEURIPS2019_01386bd6,NEURIPS2021_4ffbd5c8}. However, these equilibrium formulations are motivated primarily by memory-efficient training and amortized fixed-point solving, while how repeated latent transformations can be made more effective for fine-scale reconstruction remains a separate question.

To this end, we interpret sinusoidal recurrence as an iterative mechanism for harmonic spectral enrichment. We show that sinusoidal activations induce harmonic line spectra, providing a spectral rationale for how repeated application of a shared sinusoidal block can enrich effective spectral support without adding independently parameterized depth. In practice, this formulation achieves higher-fidelity image reconstruction with fewer parameters and optimization steps, while transferring favorably to super-resolution, NeRF, and SDF tasks.

\begin{itemize}
    \item We formulate weight-tied sinusoidal refinement for INRs, increasing effective depth through finite recurrent unrolling without adding independently parameterized layers.
    \item We provide a harmonic line-spectrum analysis of sinusoidal transformations, offering a spectral interpretation of how repeated sinusoidal refinement can enrich effective spectral support.
    \item We show that this formulation achieves higher-fidelity image reconstruction with fewer parameters and optimization steps than feed-forward baselines, while its decoder transfers favorably to continuous representation tasks.
\end{itemize}

\section{Preliminaries: Design Strategies for High-Fidelity Implicit Signal Representation}
\paragraph{\textbf{Implicit Neural Representations and Spectral Bias.}}
Implicit neural representations model signals as continuous coordinate-based functions parameterized by neural networks (e.g., $f_\theta:(x,y)\in\mathbb{R}^2\mapsto(r,g,b)\in\mathbb{R}^3$). This formulation supports continuous sampling, interpolation, and resolution-independent evaluation~\cite{10.1007/978-3-030-58452-8_24,Chen_2021_CVPR}. Despite their versatility, a central challenge is spectral bias. Coordinate MLPs tend to fit low-frequency structure before fine-scale detail~\cite{pmlr-v97-rahaman19a,ijcai2021p304}, motivating architectural choices for high-frequency representation.

\paragraph{\textbf{Enhancing Representational Capacity through Activation Functions.}}
Prior work improves high-frequency representation by modifying the activation functions of coordinate MLPs. SIREN replaces standard nonlinearities with periodic sine activations, yielding coordinate networks with strong capacity for representing complex signals~\cite{NEURIPS2020_53c04118}. FINER further adapts sinusoidal activations to cover a broader range of frequencies and improve reconstruction stability~\cite{Liu_2024_CVPR}. Beyond periodic functions, Gaussian activations provide spatially localized alternatives~\cite{10.1007/978-3-031-19827-4_9}, while WIRE employs complex Gabor wavelets that combine sinusoidal oscillation with Gaussian localization~\cite{Saragadam_2023_CVPR}.

\paragraph{\textbf{Enriching Coordinate Features with Multi-Frequency Encodings.}}
Rather than modifying the activation function, another line of work enriches the input coordinates with multi-frequency encodings. NeRF applies positional encoding to map coordinates to a multi-frequency sinusoidal basis~\cite{10.1007/978-3-030-58452-8_24}, while Random Fourier Features (RFF) use sampled sinusoidal projections to expose diverse input frequencies~\cite{NEURIPS2020_55053683}. Multiresolution encodings further augment coordinate features through learned grid or hash-based representations~\cite{10.1145/3528223.3530127}. By providing richer frequency content at the input, these methods improve the ability of INRs to fit fine-scale signals. 
\section{Latent Refinement in INRs}
\label{sec:recurrent_refinement}

Beyond modifying activations or enriching coordinate inputs, another design choice concerns how latent transformations are parameterized and evaluated across depth. We focus on this design axis through the latent transformation $\mathcal{F}$ in equation~\eqref{eq:inr_decomp}.

A feed-forward latent transformation applies a sequence of independently parameterized affine maps and nonlinearities,
\begin{equation}
    \mathbf{h}^{(\ell+1)}
    =
    \sigma_{\ell}\!\left(
        \mathbf{W}_{\ell}\mathbf{h}^{(\ell)}
        +
        \mathbf{b}_{\ell}
    \right),
    \qquad
    \ell = 0,\ldots,L-1,
    \label{eq:feedforward_refinement}
\end{equation}
where $\mathbf{h}^{(0)}=\mathcal{E}(\mathbf{x})$ and each layer has its own parameters $(\mathbf{W}_{\ell},\mathbf{b}_{\ell})$. Increasing $L$ can enrich the latent transformation, but generally requires additional independently parameterized layers.

For high-fidelity signal representation, the structure of such latent transformations can also be designed to control how frequency content is processed across depth. BACON incorporates band-limited sinusoidal filtering, while FourierNet and GaborNet construct frequency-aware feed-forward transformations~\cite{fathony2021multiplicative,Lindell_2022_CVPR}.

A separate line of work introduces iterative latent computation through an equilibrium formulation. Specifically, in equilibrium-style INR proposed by \textit{Huang}~\etal~\cite{NEURIPS2021_4ffbd5c8}, the latent representation is defined implicitly as a fixed point,
\begin{equation}
    \mathbf{h}^{\star}
    =
    \mathcal{G}_{\theta}\!\left(
        \mathbf{h}^{\star};
        \mathbf{h}^{(0)}
    \right),
    \qquad
    \mathbf{h}^{(0)}
    =
    \mathcal{E}(\mathbf{x}),
    \label{eq:equilibrium_refinement}
\end{equation}
enabling constant-memory backpropagation via implicit differentiation~\cite{NEURIPS2019_01386bd6} and amortized fixed-point solving.

Fixed-point formulations, however, characterize the latent computation only at equilibrium, making the evolution induced by successive applications of the shared map less directly visible.
We therefore consider a finite, explicitly unrolled refinement trajectory, whose intermediate states allow us to examine how repeated transformations alter the latent representation. 

\section{Harmonic Enrichment by Sinusoidal Recurrence}
\label{sec:sine_characteristics}

To analyze this stepwise refinement, we characterize the spectral structure
induced by sinusoidal transformations. We show that sinusoidal layers induce a
\emph{harmonic line-spectrum} in intermediate representations, providing a
Fourier-series interpretation of both depth and weight-tied unrolling.

\subsection{Input layer as a learnable positional embedder}
\label{sec:first_layer}
\begin{wraptable}{r}{0.60\linewidth}
  \vspace{-35pt}
  \centering
  \footnotesize
  \setlength{\tabcolsep}{3.8pt}
  \renewcommand{\arraystretch}{1.10}
  \caption{Frequency-separated reconstruction quality over training steps. Each cell reports PSNR in dB as ${\mathrm{Full}}_{\mathrm{LF}}^{\mathrm{HF}}$, where $\mathrm{Full}$ is the PSNR on the full image, $\mathrm{LF}$ is the PSNR on the low-pass component, and $\mathrm{HF}$ is the PSNR on the high-pass component.}
  \label{tab:psnr_compact_subsup_tblr}

  \resizebox{\linewidth}{!}{\begin{tblr}{
    colspec = {l | c c c c c},
    row{1}  = {font=\bfseries},
    rowsep  = 2pt,
    colsep  = 4pt,
    row{6} = {bg=OursSoft},
    row{1,2} = {abovesep=0pt, belowsep=0pt},
  }
    \toprule
    &\SetCell[c=5]{c}\#Iteration\\
    Inp. freq. ($\omega_{\rm in}$) & 200 & 400 & 600 & 800 & 1000 \\
    \midrule
    32  & ${22.73}_{37.29}^{23.51}$ & ${24.64}_{40.50}^{25.10}$ & ${25.56}_{39.01}^{26.04}$ & ${26.00}_{37.33}^{26.61}$ & ${26.46}_{37.30}^{27.09}$ \\
    64  & ${26.35}_{45.19}^{26.60}$ & ${28.91}_{48.97}^{29.05}$ & ${30.41}_{50.26}^{30.51}$ & ${31.42}_{50.10}^{31.53}$ & ${32.04}_{46.75}^{32.26}$ \\
    128 & ${30.22}_{51.84}^{30.31}$ & ${33.89}_{55.74}^{33.94}$ & ${36.11}_{56.00}^{36.18}$ & ${37.62}_{54.83}^{37.74}$ & ${38.74}_{53.98}^{38.91}$ \\
    256 & ${36.38}_{56.47}^{36.45}$ & ${40.52}_{56.45}^{40.68}$ & ${43.32}_{56.46}^{43.62}$ & ${45.39}_{56.54}^{45.87}$ & ${47.03}_{56.64}^{47.71}$ \\
    \bottomrule
  \end{tblr}}
  \vspace{-25pt}
\end{wraptable}
Following \textit{Sitzmann} \etal~\cite{NEURIPS2020_53c04118}, we define the sinusoidal coordinate lifting as
\begin{equation}
\begin{gathered}
    \mathbf{h}_0(\mathbf{x})
    \triangleq
    \mathcal{E}(\mathbf{x})\\
    =
    \sin\!\big(\omega_{\mathrm{in}}\,\mathbf{W}_{\mathrm{in}}\mathbf{x}+\mathbf{b}_{\mathrm{in}}\big),
    \label{eq:sine_encoder}
\end{gathered}
\end{equation}
where $\omega_{\mathrm{in}}$ controls the bandwidth of the embedding. Writing the $i$-th channel explicitly,
\begin{equation}
    [\mathbf{h}_0(\mathbf{x})]_i
    =
    \sin\!\big(\boldsymbol{\Omega}_i^{\!\top}\mathbf{x}+\phi_i\big),
    \qquad
    \boldsymbol{\Omega}_i \triangleq \omega_{\mathrm{in}}\mathbf{w}_i,\ \ \phi_i\triangleq b_i,
    \label{eq:sine_basis_channel}
\end{equation}
makes clear that the encoder maps the input coordinates to a collection of sinusoidal basis functions, each parameterized by a learnable frequency vector $\{\boldsymbol{\Omega}_i\}_{i=1}^{m}$ and phase $\{\phi_i\}$. Accordingly, $\mathcal{E}$ can be understood as a \emph{learnable} Fourier feature map, which is consistent with the Fourier-series interpretation of sinusoidal implicit neural representations studied by \textit{Benbarka} \etal~\cite{Benbarka_2022_WACV}. In contrast to Random Fourier Features~\cite{NEURIPS2020_55053683}, whose frequencies are fixed after initialization, the spectral components here are optimized end-to-end, while $\omega_{\mathrm{in}}$ directly determines the overall frequency scale available to the model.

\begin{figure}[t]
  \centering
  \setlength{\tabcolsep}{0pt}

  \begin{subfigure}[t]{0.18\linewidth}
    \centering
    \includegraphics[width=\linewidth]{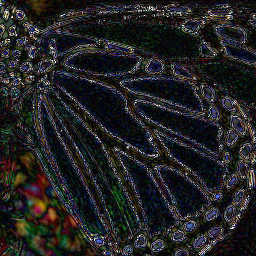}
    \caption*{\scriptsize $\omega_{\rm in}=32$}
  \end{subfigure}\hspace{0pt}%
  \begin{subfigure}[t]{0.18\linewidth}
    \centering
    \includegraphics[width=\linewidth]{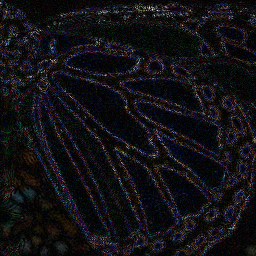}
    \caption*{\scriptsize $\omega_{\rm in}=64$}
  \end{subfigure}\hspace{0pt}%
  \begin{subfigure}[t]{0.18\linewidth}
    \centering
    \includegraphics[width=\linewidth]{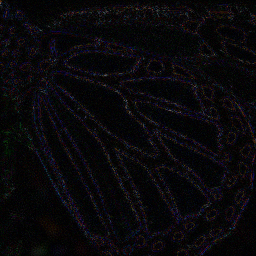}
    \caption*{\scriptsize $\omega_{\rm in}=128$}
  \end{subfigure}\hspace{0pt}%
  \begin{subfigure}[t]{0.18\linewidth}
    \centering
    \includegraphics[width=\linewidth]{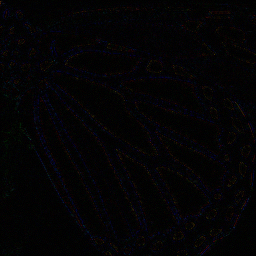}
    \caption*{\scriptsize $\omega_{\rm in}=256$}
  \end{subfigure}\hspace{0.04pt}
  \begin{subfigure}[t]{0.18\linewidth}
    \centering
    \includegraphics[width=\linewidth]{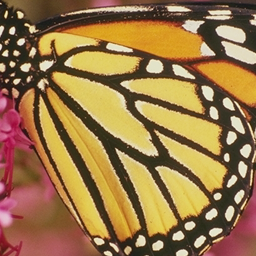}
    \caption*{\scriptsize Ground Truth}
  \end{subfigure}
  \vspace{-10pt}
\caption{\textbf{High-frequency (HF) error visualization.}
Each panel shows the absolute difference between the predicted and ground-truth \emph{high-pass} components,
$|\,\mathrm{HP}(\hat{I})-\mathrm{HP}(I_{\mathrm{GT}})\,|$, where $\mathrm{HP}(I)=I-\mathrm{GaussianBlur}(I)$.
From left to right we vary the frequency scale $\omega_{\mathrm{in}}\in\{32,64,128,256\}$ in the sinusoidal lifting (Equation~\eqref{eq:sine_encoder}), which sets the overall bandwidth of the learnable Fourier-feature embedding.
}
\vspace{-15pt}
  \label{fig:six_images_gap}
\end{figure}

This interpretation is supported empirically by the frequency-separated reconstruction results in Fig.~\ref{fig:six_images_gap} and Table~\ref{tab:psnr_compact_subsup_tblr}. As $\omega_{\mathrm{in}}$ increases, the model consistently achieves better reconstruction on the high-pass component across all training stages, while also improving full-image PSNR. A similar trend is already visible at early optimization stages (e.g., $23.51 \rightarrow 36.45$ dB at 200 iterations). 

\vspace{-10pt}
\subsection{Hidden sinusoidal layers as structured harmonic expansion}
\label{sec:harmonic_derivation}
We now show how hidden sinusoidal layers transform the encoder’s learnable sine bases into richer line spectra. Recall from Sec.~\ref{sec:first_layer} that the sinusoidal encoder produces a finite collection of basis tones:
\begin{equation}
\mathbf{h}_0(\mathbf{x})
=
\sin\!\big(\omega_{\mathrm{in}}\,\mathbf{W}_{\mathrm{in}}\mathbf{x}+\mathbf{b}_{\mathrm{in}}\big)
\in\mathbb{R}^{m},
\qquad
[\mathbf{h}_0(\mathbf{x})]_i
=
\sin\!\big(\boldsymbol{\Omega}_i^{\!\top}\mathbf{x}+\phi_i\big).
\label{eq:h0_recall}
\end{equation}
Let $\theta_i(\mathbf{x}) \triangleq \boldsymbol{\Omega}_i^{\!\top}\mathbf{x}+\phi_i$. For clarity, we first analyze the \emph{first} hidden sinusoidal layer exactly:{
\setlength{\abovedisplayskip}{5pt}
\setlength{\belowdisplayskip}{5pt}
\begin{equation}
\mathbf{h}_1(\mathbf{x})
=
\sin\!\big(\omega\,\mathbf{W}\mathbf{h}_0(\mathbf{x})+\mathbf{b}\big),
\qquad
\mathbf{W}\in\mathbb{R}^{n\times m},
\ \mathbf{b}\in\mathbb{R}^{n},
\label{eq:sine_hidden_layer}
\end{equation}}where $\sin(\cdot)$ is applied elementwise.%
\footnote{Although the formulation retains a bias term for generality, our implementation employs bias-free recurrent layers; Sec.~\ref{sec:no_bias_recurrent} provides the motivation for this choice.
} It suffices to examine one scalar pre-activation, corresponding to the $j$-th channel:
\begin{equation}
u_j(\mathbf{x})
\triangleq
[\omega\,\mathbf{W}\mathbf{h}_0(\mathbf{x})+\mathbf{b}]_j
=
\omega\,\mathbf{w}_j^{\top}\mathbf{h}_0(\mathbf{x}) + b_j,
\qquad
\mathbf{w}_j^\top := \mathbf{W}_{j:}.
\label{eq:scalar_pre_activation_exact}
\end{equation}
Substituting the encoder output from equation~\eqref{eq:h0_recall} into the pre-activation equation~\eqref{eq:scalar_pre_activation_exact}, and expanding $\mathbf{w}_j^\top\mathbf{h}_0(\mathbf{x})=\sum_{i=1}^m w_{j,i}[\mathbf{h}_0(\mathbf{x})]_i$, yields the finite multi-tone form:
\begin{equation}
u_j(\mathbf{x})
=
b_j
+
\omega\sum_{i=1}^{m} w_{j,i}\sin\!\big(\theta_i(\mathbf{x})\big)
=
\beta_j+\sum_{i=1}^{m}\alpha_{j,i}\sin\!\big(\theta_i(\mathbf{x})\big),
\label{eq:uj_exact_multitone}
\end{equation}
where $\beta_j:=b_j$ and $\alpha_{j,i}:=\omega\,w_{j,i}$. 

Applying the sinusoidal nonlinearity to this multi-tone input yields an explicit generalized Fourier expansion. Using Euler's identity ($\sin\!\big(u_j(\mathbf{x})\big) = \mathrm{Im}\!\left(e^{iu_j(\mathbf{x})}\right)$)
together with the Jacobi--Anger identity~\cite{Abramowitz_1964_HMF}, 
we obtain{
\setlength{\abovedisplayskip}{5pt}
\setlength{\belowdisplayskip}{5pt}
\begin{equation}
\exp\!\big(iu_j(\mathbf{x})\big)
=
e^{i\beta_j}
\sum_{\mathbf{k}\in\mathbb{Z}^{m}}
\Big(\prod_{i=1}^{m} J_{k_i}(\alpha_{j,i})\Big)\,
\exp\!\Big(i\sum_{i=1}^{m} k_i\theta_i(\mathbf{x})\Big),
\label{eq:multi_jacobi_anger_exact}
\end{equation}}and therefore{
\setlength{\abovedisplayskip}{5pt}
\setlength{\belowdisplayskip}{5pt}
\begin{equation}
\sin\!\big(u_j(\mathbf{x})\big)
=
\sum_{\mathbf{k}\in\mathbb{Z}^{m}}
c_{j,\mathbf{k}}\,
\sin\!\Big(
\beta_j+\sum_{i=1}^{m} k_i\theta_i(\mathbf{x})
\Big),
\qquad
c_{j,\mathbf{k}}
=
\prod_{i=1}^{m} J_{k_i}(\alpha_{j,i}).
\label{eq:explicit_fourier_sine_multitone_exact}
\end{equation}}Equation~\eqref{eq:explicit_fourier_sine_multitone_exact} makes the sinusoid-specific effect explicit: in other words, a hidden sine layer creates new spectral lines at integer combinations of the encoder frequencies---including sums, differences, and higher-order harmonics---rather than merely reweighting existing ones~\cite{roddenberry2024implicit}. These frequencies take the form{
\setlength{\abovedisplayskip}{5pt}
\setlength{\belowdisplayskip}{5pt}
\begin{equation}
\boldsymbol{\Omega}'
=
\sum_{i=1}^{m} k_i\,\boldsymbol{\Omega}_i,
\qquad
\mathbf{k}\in\mathbb{Z}^{m},
\label{eq:integer_combination_freqs_exact}
\end{equation}}that is, integer combinations of the encoder frequencies, as also characterized in the harmonic analysis of sinusoidal networks by \textit{Novello}~\etal~\cite{Novello_2025_CVPR}.

This exact calculation motivates the deeper-layer view. Once a hidden representation is itself regarded as a line spectrum, or a finite truncation of its dominant tones, applying another sinusoidal layer yields the same type of integer-combination closure. Thus, if $\Omega^{(\ell)}$ denotes the set of prominent frequency vectors after layer $\ell$, then in this idealized sense{
\setlength{\abovedisplayskip}{5pt}
\setlength{\belowdisplayskip}{5pt}
\begin{equation}
\Omega^{(\ell+1)}
\subseteq
\mathrm{span}_{\mathbb{Z}}\!\big(\Omega^{(\ell)}\big).
\label{eq:support_growth_depth}
\end{equation}}

\begin{wraptable}{r}{0.58\linewidth}
  \vspace{-35pt}
  \centering
  \footnotesize
  \caption{Reconstruction quality at optimization step $500$ as a function of the number of recurrent unrolling steps. `Feed-forward' corresponds to a single pass, and \#2--\#5 denote recurrent steps $R{=}2$--$5$. Each entry is formatted as ${\textbf{PSNR}}_{\Delta}$, where $\Delta$ is the PSNR gain (dB) relative to the feed-forward baseline.}
  \label{tab:psnr_recurrent_tblr}

  \resizebox{\linewidth}{!}{%
  \begin{tblr}{
    colspec = {l c c c c c},
    row{1,3}  = {font=\bfseries},
    rowsep  = 2pt,
    colsep  = 4pt,
    column{6} = {bg=OursSoft},
    row{1,2} = {abovesep=0pt, belowsep=0pt},
    row{3,4} = {abovesep=0.5pt, belowsep=0.5pt},
    row{4}  = {fg=gray},
  }
    \toprule
    &\SetCell[c=5]{c}\#Recurrent Unrolling Steps\\
    &Feed-forward & \#2 & \#3 & \#4 & \#5 \\
    \midrule
    PSNR&${39.724}_{+0.0}$ & ${46.362}_{+6.6}$ & ${54.772}_{+15.0}$ & ${60.843}_{+21.1}$ & $\textbf{63.378}_{+23.7}$ \\
    \#Param.&593.7K&593.7K&593.7K&593.7K&593.7K\\
    \bottomrule
  \end{tblr}}
  \vspace{-25pt}
\end{wraptable}

\noindent\!Hence, increasing depth enables progressively higher-order harmonic interactions from the same coordinate lifting. The same viewpoint also motivates weight-tied unrolling: repeatedly applying a shared sinusoidal block realizes iterative harmonic refinement under a fixed parameter budget.

Table~\ref{tab:psnr_recurrent_tblr} quantifies the practical effect of finite recurrent refinement under a fixed parameter budget. At optimization step $500$, increasing the number of recurrent unrolling steps from a single feed-forward pass to $R=5$ improves PSNR from $39.724$ dB to $63.378$ dB, with monotonic gains at every intermediate step under the same parameter count. We next examine whether this improvement is accompanied by systematic changes in the intermediate spectral structure.

On a regular $H\times H$ coordinate grid, we analyze the intermediate feature maps $\mathbf{h}^{(\ell)}$ by computing the per-channel 2-D DFT $\mathbf{F}^{(\ell)}_c{=}\mathrm{DFT}(\mathbf{h}^{(\ell)}_{:,c})$, where $|\mathbf{F}^{(\ell)}_c[u,v]|$ is the strength of channel $c$ at frequency $(u,v)$. After channel averaging and peak normalization in dB,
{
\setlength{\abovedisplayskip}{0pt}\setlength{\belowdisplayskip}{2pt}
\begin{equation}
M^{\mathrm{dB}}_{\ell}[u,v]=20\log_{10}\!\frac{(1/C)\sum_c|\mathbf{F}^{(\ell)}_c[u,v]|}{\max_{u',v'}(\,\cdot\,)},
\label{eq:db}
\end{equation}}we summarize each refinement step using spectral support $S_\ell^\tau$ (which measures spectral breadth), and DC-normalized upper-band spectral content $\mathrm{HF}_\ell$ (which measures the average spectral magnitude over the upper radial band $\mathcal{U}{=}\{H/4,\ldots,H/2{-}1\}$ relative to the DC component):
{
\setlength{\abovedisplayskip}{3pt}
\setlength{\belowdisplayskip}{3pt}
\begin{equation}
S_\ell^{\tau}\triangleq H^{-2}\big|\{(u,v):M^{\mathrm{dB}}_\ell\!>\!\tau\}\big|,\quad
\mathrm{HF}_\ell\triangleq\tfrac{1}{|\mathcal{U}|}\!\sum_{k\in\mathcal{U}}\tfrac{R_\ell(k)}{R_\ell(0)},
\label{eq:hf}
\end{equation}}with $R_\ell(k)$ denoting the radial profile of the channel-averaged linear magnitude spectrum at radius $k$ from the zero-frequency (DC) component.

At random initialization, repeated sinusoidal transformations rapidly expand the measured spectral support and upper-band content across refinement steps (Table~\ref{tab:spec}(a)). After training, its trajectory reflects task-dependent reweighting. In contrast, the equilibrium-style iSIREN baseline~\cite{NEURIPS2021_4ffbd5c8} exhibits a stationary spectral profile, while non-sinusoidal recurrent controls lose support (Table~\ref{tab:spec}(b,c)). This distinction is mirrored in fitting: under matched capacity, recurrence consistently improves SIREN~\cite{NEURIPS2020_53c04118} and benefits FINER~\cite{Liu_2024_CVPR} beyond $\sim$200K parameters, but degrades Gaussian models and yields no persistent gain for PEMLP (Table~\ref{tab:recon_quality}). 

\begin{table}[t!]
\caption{\textbf{Spectral support $S_\ell^\tau$ and radial HF$_\ell$.}}
\label{tab:spec}
\vspace{-10pt}
\centering
\footnotesize
\setlength{\tabcolsep}{3pt}
\resizebox{\linewidth}{!}{%
\begin{tblr}{
      width = \linewidth,
      colspec = {@{} X[l] c c | *{9}{c} | c c | c @{}},
      row{1} = {
        abovesep = .5pt,
        belowsep = .5pt,
      },
      abovesep = 0pt,
      belowsep = 0pt,
      row{3,6} = {bg=OursSoft},
      row{9} = {bg=OursSoft!70},
      row{14} = {bg=LightBeige},
    }
\toprule
\SetCell[r=2]{l}\textbf{Architecture} & \SetCell[r=2]{c}\textbf{Recurrent} & \SetCell[r=2]{c}\textbf{Mode}
  & \SetCell[c=9]{c} \textbf{$S_\ell^\tau$ per\ depth/recurrent $\ell$ (\%)} & & & & & & & & &
  \SetCell[c=2]{c} \textbf{HF$_\ell$ at $\ell{=}0,8$} & &
  \textbf{PSNR} \\
 & &
  & \textbf{0} & \textbf{1} & \textbf{2} & \textbf{3} & \textbf{4} & \textbf{5} & \textbf{6} & \textbf{7} & \textbf{8}
  & \textbf{0} & \textbf{8}
  & \textbf{\textit{for ref.}} \\
\midrule
\SetCell[c=15]{l} \textit{(a) Sinusoidal recurrent, untrained} & & & & & & & & & & & & & & \\
SIREN      & \cmark & $\omega{=}256/45$ & 30.9 & 26.5 & 34.3 & 43.7 & 52.0 & 61.3 & 72.3 & 85.0 & \textbf{94.6} & 0.022 & 0.101 & ---  \\
FINER          & \cmark & $\omega{=}256/45$ & 51.4 & 53.6 & 72.4 & 89.3 & \textbf{100}  & 100  & 100  & 100  & 100           & 0.048 & 0.232 & ---  \\
\SetCell[c=15]{l} \textit{(b) Sinusoidal recurrent, trained on Set 5} & & & & & & & & & & & & & & \\
SIREN      & \cmark & Float               & 31.0 & 17.1 & 17.2 & 19.1 & 20.5 & 12.9 & 16.4 & 28.1 & 61.7          & 0.020 & 0.011 & 78.35 \\
FINER          & \cmark & Float               & 53.9 & 41.4 & 46.6 & 46.0 & 37.4 & 16.4 & 41.6 & 100  & 100           & 0.044 & 0.016 & 85.62 \\
\SetCell[c=15]{l} \textit{$\quad\blacktriangleright$ Sinusoidal-based Spectral methods} & & & & & & & & & & & & & & \\
iSIREN~\cite{NEURIPS2021_4ffbd5c8}& \cmark & DEQ& 32.4 & 33.5 & 33.5 & 33.5 & 33.5 & 33.5 & 33.5 & 33.5 & 33.5& 0.008 & 0.008 & 47.16 \\
GaborNet~\cite{fathony2021multiplicative}   & \xmark & $L{=}3$   & 40.6 & 58.2 & 73.0 & 35.2          & --- & --- & --- & --- & --- & 0.038 & 0.008 & 44.48 \\
FourierNet~\cite{fathony2021multiplicative} & \xmark & $L{=}3$    & 41.4 & 58.3 & 66.6 & 36.0          & --- & --- & --- & --- & --- & 0.050 & 0.008 & 38.38 \\
BACON~\cite{Lindell_2022_CVPR} & \xmark & freq$=$128 & 0.9  & 6.3  & 14.0 & 15.1          & --- & --- & --- & --- & --- & 0.059 & 3e-4  & 27.88 \\
\SetCell[c=15]{l} \textit{(c) Non-sinusoidal recurrent, trained --- \textcolor{red}{\textbf{spectrum collapses}}} & & & & & & & & & & & & & & \\
Gauss          & \cmark & $\sigma{=}30$     & 1.0  & 1.9  & 14.0 & 1.2  & \textbf{0.0} & 0.0 & 0.0          & 0.0 & 0.0 & 2e-4 & 7e-3 & 12.02 \\
PEMLP         & \cmark & $N_f{=}10$        & 0.8  & 0.9  & 0.8  & 0.5  & 0.1          & 0.1 & \textbf{0.0} & 0.0 & 0.0 & 6e-4 & 2e-5 & 24.09 \\
\bottomrule
\end{tblr}
}
\vspace{-10pt}
\vspace{-3pt}
\end{table}
\begin{table*}[t!]
\vspace{4pt}
\caption{\textbf{Effectiveness of Recurrent Connections in Non-sinusoidal and Sinusoidal Architectures.} PSNR (dB)$\uparrow$ is reported in the capacity-limited regime.}
\label{tab:recon_quality}
\vspace{-12pt}
\centering
\scriptsize

\begin{subtable}[t]{0.49\linewidth}
\centering
\resizebox{\linewidth}{!}{%
\begin{tblr}{
    colspec = {@{} l c *{4}{c} @{}},
    row{1} = {
        font=\bfseries,
        abovesep=1pt,
        belowsep=1pt,
    },
    rowsep=1pt,
    colsep=2pt,
    abovesep=1pt,
    belowsep=1pt,
    row{3,5} = {bg=OursSoft},
}
\toprule
\textbf{Non-sinusoidal} & \textbf{Rec.}
& \textbf{$\sim$100K}
& \textbf{$\sim$200K}
& \textbf{$\sim$400K}
& \textbf{$\sim$600K} \\
\midrule
PEMLP (1e-3) & \xmark
& \textbf{28.89} & \textbf{31.04} & 31.80 & \textbf{34.20} \\
+ Rec. (1e-3) & \cmark
& 27.48 & 29.64 & \textbf{33.82} & 31.04 \\
Gauss (1.5e-4) & \xmark
& \textbf{20.48} & \textbf{26.49} & \textbf{35.33} & \textbf{60.31} \\
+ Rec. (1.5e-4) & \cmark
& 11.74 & 11.77 & 11.92 & 12.02 \\
\bottomrule
\end{tblr}%
}
\end{subtable}
\hfill
\begin{subtable}[t]{0.49\linewidth}
\centering
\resizebox{\linewidth}{!}{%
\begin{tblr}{
    colspec = {@{} l c *{4}{c} @{}},
    row{1} = {
        font=\bfseries,
        abovesep=1pt,
        belowsep=1pt,
    },
    rowsep=1pt,
    colsep=2pt,
    abovesep=1pt,
    belowsep=1pt,
    row{3,5} = {bg=OursSoft},
}
\toprule
\textbf{Sinusoidal} & \textbf{Rec.}
& \textbf{$\sim$100K}
& \textbf{$\sim$200K}
& \textbf{$\sim$400K}
& \textbf{$\sim$600K} \\
\midrule
SIREN (1e-3) & \xmark
& 33.60 & 39.57 & 46.23 & 48.64 \\
+ Rec. (1.5e-4) & \cmark
& \textbf{37.26} & \textbf{46.26} & \textbf{61.36} & \textbf{78.35} \\
FINER (5e-4) & \xmark
& \textbf{38.01} & 40.59 & 45.56 & 48.36 \\
+ Rec. (1.5e-4) & \cmark
& 34.21 & \textbf{44.45} & \textbf{68.06} & \textbf{85.62} \\
\bottomrule
\end{tblr}%
}
\end{subtable}

\vspace{-15pt}
\end{table*}

\vspace{-10pt}
\subsection{On the Use of Bias Terms in Recurrent Layer}
\label{sec:no_bias_recurrent}
To examine the role of bias in weight-tied sinusoidal recurrence, we first consider the general recurrent update with an additive bias term. Let{
\setlength{\abovedisplayskip}{3pt}
\setlength{\belowdisplayskip}{3pt}
\begin{equation}
\mathbf h^{(r+1)}
=
\sin\left(
\omega\big(
\mathbf W_{\rm rec}\mathbf h^{(r)}
+
\mathbf b_{\rm rec}
\big)
\right),
\quad
\mathbf{z}^{(r)}
\triangleq
\omega \mathbf{W}_{\rm rec}\mathbf{h}^{(r)},
\quad
\boldsymbol{\beta}
\triangleq
\omega\mathbf{b}_{\rm rec}.
\end{equation}}By applying the sine addition formula element-wise across channels, the recurrent update can be written as{
\setlength{\abovedisplayskip}{3pt}
\setlength{\belowdisplayskip}{3pt}
\begin{equation}
\begin{gathered}
    \mathbf{h}^{(r+1)}
    =
    \sin\!\big(
        \mathbf{z}^{(r)}+\boldsymbol{\beta}
    \big)
    =
    \sin\!\big(
        \mathbf{z}^{(r)}
    \big)
    \odot
    \cos(\boldsymbol{\beta})
    +
    \cos\!\big(
        \mathbf{z}^{(r)}
    \big)
    \odot
    \sin(\boldsymbol{\beta}).
\end{gathered}
    \label{eq:recurrent_bias_phase_shift}
\end{equation}}Here, $\mathbf{b}_{\rm rec}$ induces the same phase shift whenever the tied nonlinear map is reused. This shift changes the phase-dependent derivative gate through which variations in the current hidden state are propagated to the next step,
\begin{equation}
    \frac{
        \partial\mathbf{h}^{(r+1)}
    }{
        \partial\mathbf{h}^{(r)}
    }
    =
    \operatorname{diag}\!\left[
        \cos\!\big(
            \mathbf{z}^{(r)}+\boldsymbol{\beta}
        \big)
    \right]
    \omega\mathbf{W}_{\rm rec}.
    \label{eq:recurrent_bias_jacobian}
\end{equation}

\begin{wrapfigure}{r}{0.50\textwidth}
\vspace{-35pt}
\centering
\footnotesize

\begingroup
\captionof{table}{Mean peak PSNR (dB) over 1K iterations on Kodak24. We use $\omega_0=256/45$ (first/hidden layers). OFF and ON denote recurrent bias disabled and enabled, respectively. Both settings use 792K parameters.}
\label{tab:psnr-recurrence-depth}

\resizebox{\linewidth}{!}{%
\begin{tblr}{
    colspec = {l c c c c c},
    row{3} = {font=\bfseries},
    rowsep = 2pt,
    colsep = 4pt,
    row{1} = {abovesep=0pt, belowsep=0pt},
    row{2,3,4} = {abovesep=0.5pt, belowsep=0.5pt},
    row{4} = {fg=gray},
    row{3} = {bg=OursSoft},
    row{2} = {bg=LightBeige},
}
    \toprule
    \#Rec. $\to$ & $1$ & $2$ & $3$ & $4$ & $5$ \\
    \midrule
    Bias ON  & 51.16 & 60.35 & 51.97 & 63.57 & 78.10 \\
    Bias OFF & 51.41 & 83.49 & 87.54 & 89.59 & \textbf{90.28} \\
    $\Delta_{\rm PSNR}$ & +0.25 & \textbf{+23.14} & \textbf{+35.57} & \textbf{+26.02} & \textbf{+12.18} \\
    \bottomrule
\end{tblr}}
\endgroup

\captionsetup[subfigure]{skip=0pt}

\begin{subfigure}[b]{0.49\linewidth}
    \centering
    \includegraphics[width=\linewidth]{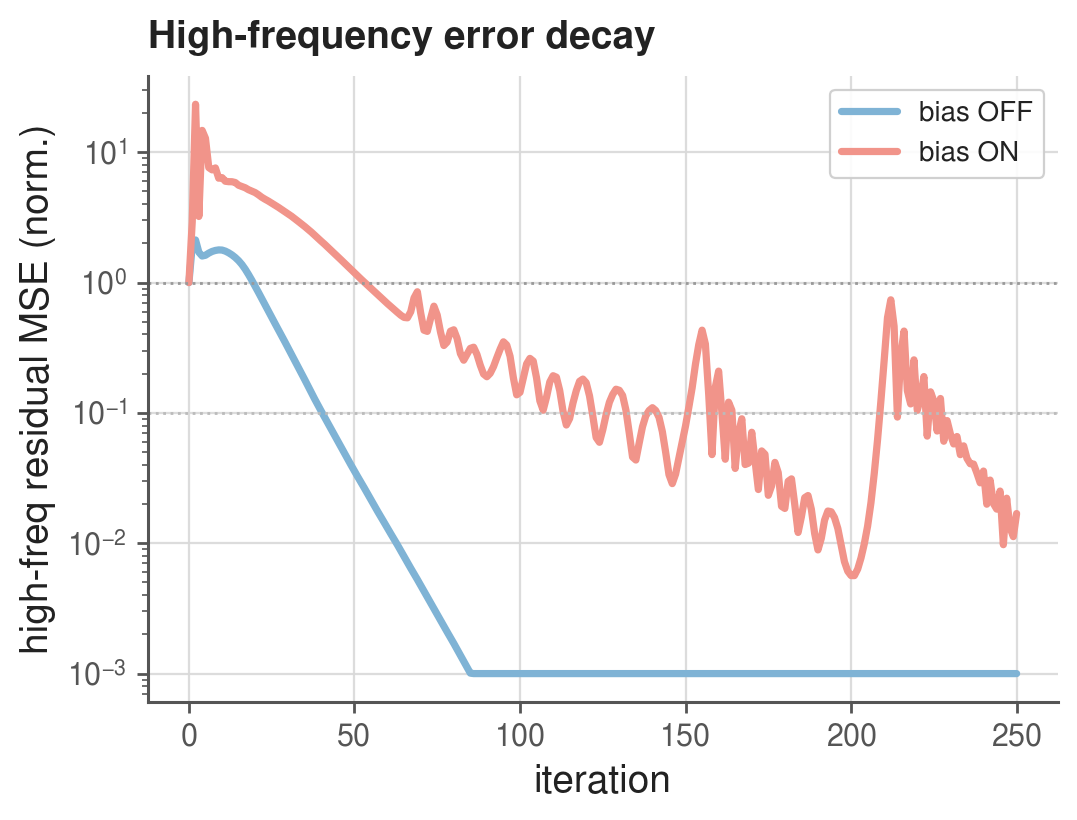}
    \subcaption{MSE Error.}
    \label{fig:highfreq_decay}
\end{subfigure}\hfill
\begin{subfigure}[b]{0.49\linewidth}
    \centering
    \includegraphics[width=\linewidth]{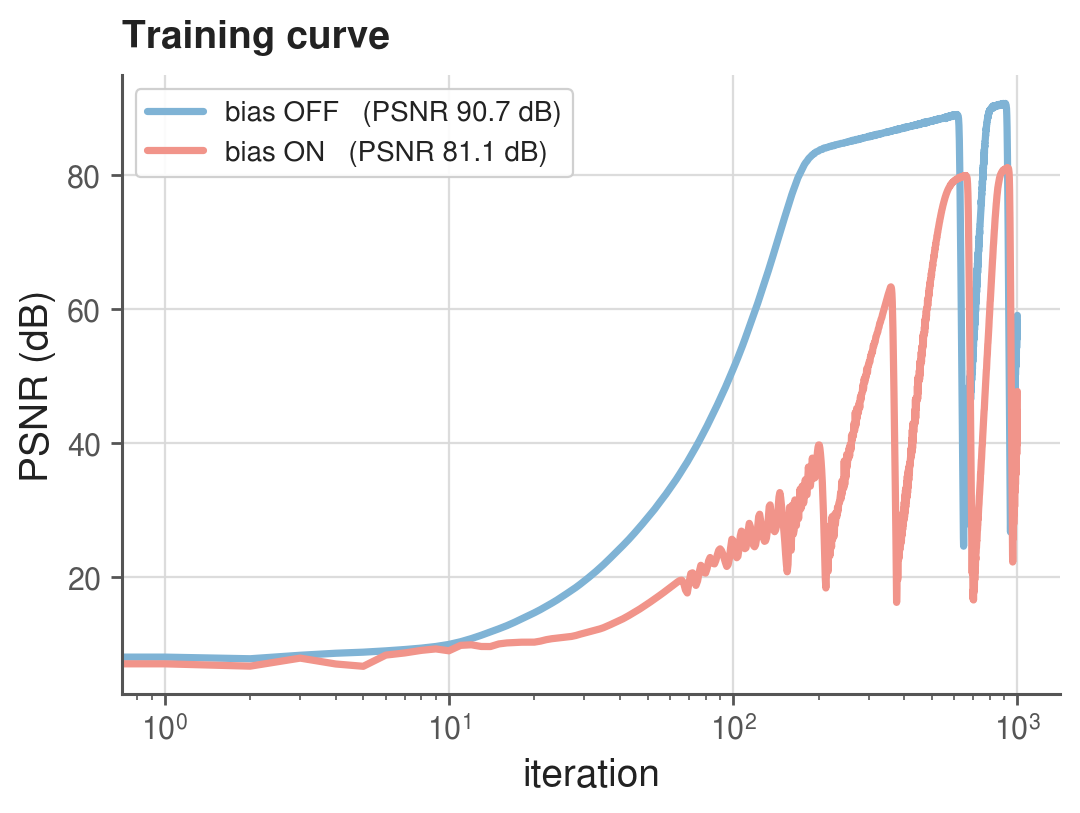}
    \subcaption{Training curve.}
    \label{fig:training_curve}
\end{subfigure}

\vspace{-8pt}
\captionof{figure}{Effectiveness of $\mathbf b_{\rm rec}$ for training trajectory.}
\label{fig:decay_training}
\vspace{-20pt}
\end{wrapfigure}

\noindent{That is, $\boldsymbol{\beta}$ shifts each channel to a different operating point on the sinusoidal response, replacing the bias-free local gate $\cos(\mathbf{z}^{(r)})$ with $\cos(\mathbf{z}^{(r)}+\boldsymbol{\beta})$. Because this modulation is reapplied at every unrolled step, its effect can accumulate along the recurrent trajectory. Empirically, models with and without $\mathbf{b}_{\rm rec}$ behave similarly for a single recurrent application ($R=1$), whereas enabling the recurrent bias leads to a substantial degradation in fitting once the same map is reused across multiple steps ($R\geq2$; Table~\ref{tab:psnr-recurrence-depth}). We further observe less stable high-frequency residual dynamics when recurrent bias is enabled (Fig.~\ref{fig:decay_training}). }

\section{Method}
\label{sec:method}
\begin{figure}[t!]
    \centering
    \includegraphics[width=.865\linewidth]{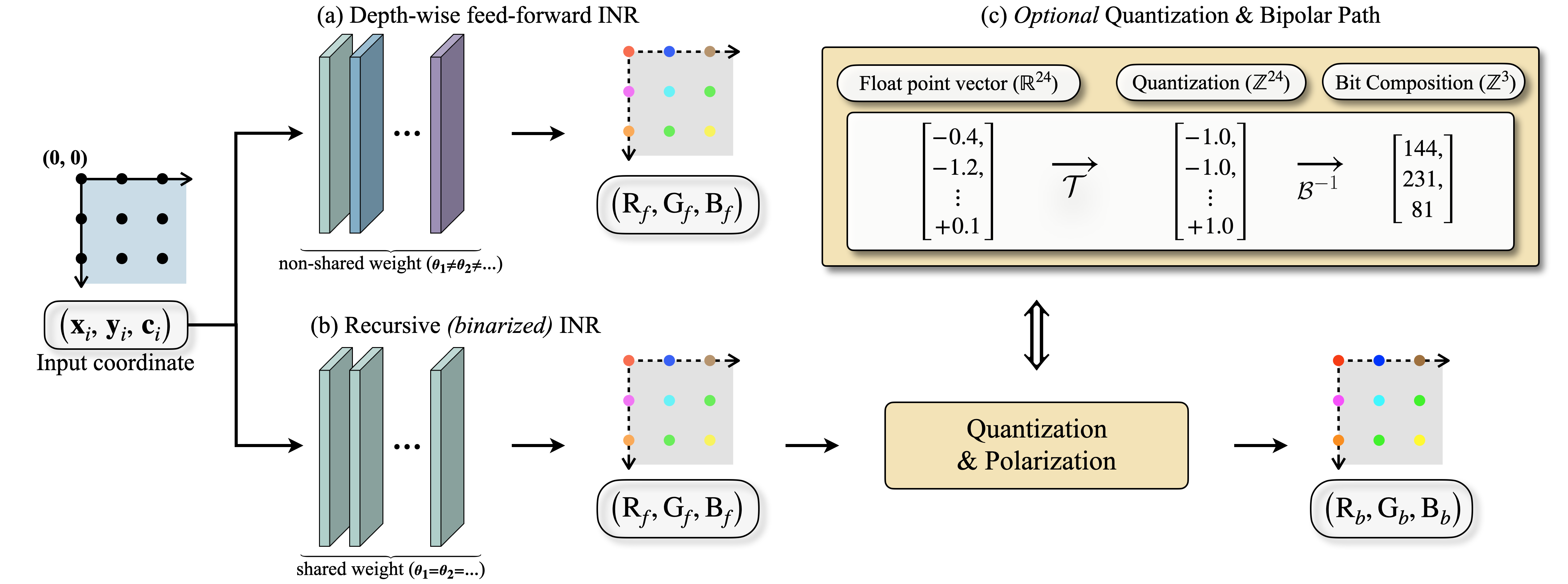}
    \vspace{-5pt}
    \caption{\textbf{Feed-forward vs. recurrent INRs under a fixed parameter budget.} (a) A depthwise feed-forward INR baseline uses non-shared weights across layers to map coordinates ($x_i, y_i$) to RGB outputs ($R_f, G_f, B_f$) in FP32. (b) recurrent INR reuses a single shared module and iteratively refines hidden features for $T$ steps, increasing effective depth via unrolling and training with backpropagation through time. Optionally, we apply a threshold/sign operator $\mathcal{T}$ to obtain bipolar outputs and an inverse bipolar mapping $\mathcal{B}^{-1}$ to convert bipolar values to UINT8 for exact discrete recovery.}
    \label{fig:figure1}
    \vspace{-15pt}
\end{figure}

As discussed in Sec.~\ref{sec:sine_characteristics}, repeated sinusoidal transformations progressively enrich the reachable spectrum through structured harmonic interactions. This observation motivates a recurrent, weight-tied design: instead of increasing the number of independently parameterized layers, we repeatedly apply a shared sinusoidal block to iteratively refine the latent representation under a fixed parameter budget. The overall pipeline is illustrated in Fig.~\ref{fig:figure1}.

Given a coordinate $\boldsymbol{x}$, we initialize a latent state by a sinusoidal input projection and refine it for $R$ recurrent steps using a shared transformation:
\begin{equation}
\mathbf{h}^{(0)} = \sigma\big(W^{(\mathrm{in})}(\boldsymbol{x})\big), \qquad
\mathbf{h}^{(r)} = \sigma\big(W^{(\mathrm{rec})}(\mathbf{h}^{(r-1)})\big), \quad r=1,\ldots,R,
\label{eq:recurrent_update}
\end{equation}
followed by the output projection
\begin{equation}
\hat{\mathbf{c}} = W^{(\mathrm{out})}(\mathbf{h}^{(R)}).
\label{eq:model_arch_recursive}
\end{equation}
Here, $W^{(\mathrm{in})}$, $W^{(\mathrm{rec})}$, and $W^{(\mathrm{out})}$ are bias-free linear maps, with $W^{(\mathrm{rec})}$ shared across all recurrent steps, and $\sigma(\cdot)$ denotes the elementwise sinusoidal activation. Equivalently, the model can be written in compact compositional form as
\begin{equation}
f_\psi(\boldsymbol{x})
=
W^{(\mathrm{out})}
\circ
\underbrace{
\big(\sigma \circ W^{(\mathrm{rec})}\big)\circ \cdots \circ \big(\sigma \circ W^{(\mathrm{rec})}\big)
}_{R\ \text{recurrent steps}}
\;\circ\;
\sigma \circ W^{(\mathrm{in})}
(\boldsymbol{x}).
\label{eq:model_arch}
\end{equation}
This recurrence is the architectural counterpart of the spectral analysis in Sec.~\ref{sec:sine_characteristics}: each application of the shared sinusoidal block expands harmonic interactions while preserving the parameter count.

\subsection{Binarized supervision for exact discrete reconstruction}
For lossless representation, we seek to reproduce the quantized values observed at sampled coordinates exactly, rather than merely approximate them through real-valued regression. This motivates supervising the INR in a binarized code space instead of directly regressing the corresponding intensity or amplitude values. In this sense, our formulation is related to recent INR-based approaches that cast reconstruction in a digital representation space~\cite{Han_2025_CVPR}.

Let $y_p \in \{0,\ldots,2^B-1\}$ be the quantized target value at coordinate $p$, and let $\Gamma(\cdot)$ be a $B$-bit encoder. We define the supervision target as a bipolar code
\begin{equation}
\mathbf{c}_p
=
2\Gamma(y_p)-1
\in
\{-1,+1\}^{B}.
\label{eq:bipolar_code}
\end{equation}
We use a bipolar code rather than a standard $\{0,1\}$ code for two reasons. First, bipolar targets are zero-centered, which is better matched to the symmetric range of sinusoidal activations. Second, all bipolar codewords have identical norm, i.e., $\|\mathbf{c}_p\|_2^2 = B$, which yields a particularly simple alignment-based training objective. Specifically, the network predicts a real-valued code vector $\hat{\mathbf{c}}_p = f_\psi(\boldsymbol{x}_p) \in \mathbb{R}^{B}$, and we optimize it using cosine similarity:
\begin{equation}
\max_{\psi}\;\; \mathcal{L}_{\mathrm{align}}(\psi)
\quad\text{with}\quad
\mathcal{L}_{\mathrm{align}}
=
\frac{1}{|\mathcal{P}|}
\sum_{p\in\mathcal{P}}
\frac{\hat{\mathbf{c}}_p^{\top}\mathbf{c}_p}
{\|\hat{\mathbf{c}}_p\|_2\,\|\mathbf{c}_p\|_2},
\qquad
\hat{\mathbf{c}}_p=f_\psi(\boldsymbol{x}_p).
\label{eq:cosine_loss_min}
\end{equation}
where $\mathcal{P}$ denotes the set of sampled coordinates. This choice is also well motivated from the squared-error perspective. Expanding the Euclidean distance between prediction and target gives
\begin{equation}
\|\hat{\mathbf{c}}_p-\mathbf{c}_p\|_2^2
=
\|\hat{\mathbf{c}}_p\|_2^2
+
\|\mathbf{c}_p\|_2^2
-
2\hat{\mathbf{c}}_p^{\top}\mathbf{c}_p.
\label{eq:mse_expand}
\end{equation}
Since $\mathbf{c}_p \in \{-1,+1\}^B$, we have $\|\mathbf{c}_p\|_2^2 = B$, which is constant across all targets. Therefore, the target-dependent term is entirely determined by the alignment between $\hat{\mathbf{c}}_p$ and $\mathbf{c}_p$. Cosine similarity makes this alignment explicit while additionally removing sensitivity to the scale of $\hat{\mathbf{c}}_p$.

\paragraph{Gray-binarization.}
In our implementation, $\Gamma(\cdot)$ is chosen as a Gray encoder~\cite{gray1953pulse}. Compared with standard binary coding, Gray coding preserves local adjacency in the quantized signal space, since neighboring quantization levels differ by only one bit. We keep this design choice simple in the method description and return to its implications in the Sec.~\ref{sec:discussion}. 

\tikzset{
    largewindow_w/.style={black, line width=0.30mm},
    smallwindow_w/.style={black, line width=0.10mm},
    closeup_w_3/.style={
        opacity=1.0,
        height=1.cm,
        width=1.cm,
        connect spies,
        black
    },
}

\newcommand{\SirenImage}[5]{%
    \begin{tikzpicture}[x=5cm, y=5cm, spy using outlines={every spy on node/.append style={smallwindow_w}}]
        \node[anchor=south, inner sep=0pt] (FigA) at (0,0) {
            \includegraphics[trim=0 0 0 0, clip, width=0.84in]{#1}
        };
        
        \spy [closeup_w_3, magnification=3, size=1.0cm] on ($(FigA)+(-0.165,-0.165)$)
            in node[largewindow_w, anchor=east, size=1.3cm] at ($(FigA.north)+(0.18,-0.25)$);
        
        \node[
            fill=white, 
            fill opacity=0.85,  
            text opacity=1,     
            draw=none,          
            rounded corners=2pt,
            minimum width=0.8cm, 
            inner sep=2pt,
            anchor=center, 
            font=\sffamily\scriptsize\bfseries 
        ] at ($(FigA.south west)+(#4,#5)$) {#2};
        
        \node[
            anchor=north, 
            font=\sffamily\scriptsize 
        ] at ($(FigA.south)+(0, -0.01)$) {#3};
    \end{tikzpicture}
}

\begin{figure*}[t]
    \centering
    \vspace{-5pt}
    \SirenImage{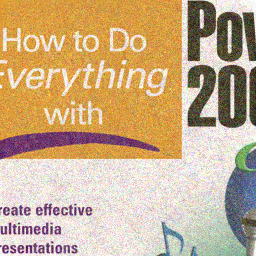}{17.85}{Gauss}{0.100}{0.385}
    \hspace{-3.5mm}
    \SirenImage{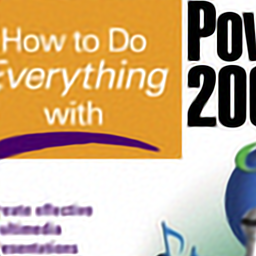}{23.52}{FINER}{0.100}{0.385}
    \hspace{-3.5mm}
    \SirenImage{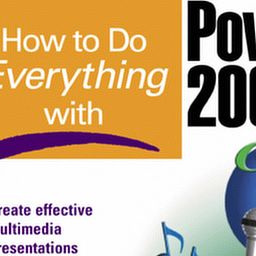}{32.88}{FourierNet}{0.100}{0.385}
    \hspace{-3.5mm}
    \SirenImage{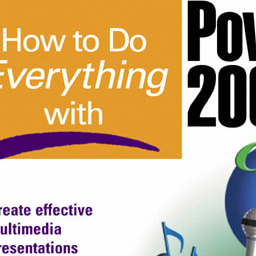}{36.60}{iSIREN}{0.100}{0.385}
    \hspace{-3.5mm}
    \SirenImage{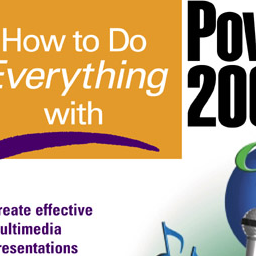}{Zero Bit Error}{\textbf{OURS}}{0.220}{0.385}
    \hspace{-3.5mm}
    \vspace{-10pt}
\caption{\textbf{Qualitative comparison after 125 optimization iterations} on an early-converging example that reaches zero bit error at the 125-iteration checkpoint. The boxes show PSNR values, and inset crops visualize fine-detail recovery. Compared to baselines, our method reconstructs sharper edges and text-like structures with fewer artifacts, consistent with the quantitative gains in Table~\ref{tab:main_result_clean}.}
    \label{fig:main_qual}
    \vspace{-18pt}
\end{figure*}
\section{Experiments}
\label{sec:experiments}

We first evaluate quantized 2D image fitting, where exactness is directly measurable under a fixed parameter budget. We then assess transferability by integrating the same decoder into super-resolution and 3D NeRF pipelines.

\subsection{Image representation}
\paragraph{\textbf{Datasets and Evaluation Protocol.}}
We evaluate on four widely used image datasets: Set5~\cite{BMVC_26_135}, Kodak24~\cite{Kodakdataset}, DIV2K~\cite{DIV2kdataset}, and FFHQ~\cite{Karras_2019_CVPR}. For DIV2K and FFHQ, we randomly sample 100 and 600 images, respectively, and resize all images to $256 \times 256$. Unless otherwise noted, we use five recurrent steps. For fair comparison, we preserve the original macro-architecture of the baselines and adjust only the hidden width to obtain comparable parameter counts.

\paragraph{\textbf{Implementation.}}
The decoder output is bounded by $\mathcal{G}(z)=\sin(\arctan z)$, which can be written as $z/\sqrt{1+z^2}$. This function is approximately linear around the origin while smoothly saturating to $[-1,1]$, making it well matched to the bipolar target space and empirically stabilizing optimization. To keep training fully differentiable, the transformations $\mathcal{T}$ and $\mathcal{B}^{-1}$ are applied only at inference time. We train all models with AdamW using a learning rate of $1.5 \times 10^{-4}$, without a scheduler, on NVIDIA RTX 3090 GPUs.

\begin{table*}[t!]
\centering
\footnotesize
\caption{We report the number of optimization iterations, reconstruction fidelity, and bit error, along with parameter count. Baselines are trained for 1000 iterations with 791K parameters. Our model uses fewer parameters (609K) and achieves substantially higher fidelity with a fixed budget of 100 iterations (OURS (\#iter=100)), and can reach exact quantized reconstruction with an adaptive number of iterations.
}
\vspace{-10pt}
\resizebox{\linewidth}{!}{%
\begin{tblr}{
width=\textwidth,
colspec={l c c | c c | c c | c c | c c},
row{1-2} = {font=\bfseries},
row{3} = {bg=LightBeige},
row{8,10} = {bg=PeachBeige},
row{14} = {bg=OursSoft},
abovesep=2pt, belowsep=2pt,
row{3,8,10,14} = {abovesep=0pt, belowsep=0pt}, 
}
\toprule
\SetCell[r=2]{c}\textbf{Method}
& \SetCell[r=2]{c}\textbf{\#Param.}
& \SetCell[r=2]{c}\textbf{\#Iter}
& \SetCell[c=2]{c}Set5
&
& \SetCell[c=2]{c}Kodak24
&
& \SetCell[c=2]{c}DIV2K-100
&
& \SetCell[c=2]{c}FFHQ-600
& \\
&
&
& PSNR / SSIM $\uparrow$
& Bit err $\downarrow$
& PSNR / SSIM $\uparrow$
& Bit err $\downarrow$
& PSNR / SSIM $\uparrow$
& Bit err $\downarrow$
& PSNR / SSIM $\uparrow$
& Bit err $\downarrow$ \\
\midrule
\SetCell[c=11]{l}Standard INR Baselines \\
$\circ$ WIRE
& \SetCell[r=4]{c}\textbf{791K}
& \SetCell[r=4]{c}\textbf{1000}
& 26.91 / .6637 & 544.52K
& 26.83 / .6583 & 566.38K
& 26.65 / .7620 & 555.53K
& 26.43 / .5971 & 561.66K \\
$\circ$ Gauss
& 
& 
& 33.97 / .8802 & 453.77K
& 34.18 / .8867 & 460.84K
& 34.06 / .9211 & 456.13K
& 34.01 / .8647 & 458.33K \\
$\circ$ SIREN
& 
& 
& 38.54 / .9629 & 350.11K
& 35.01 / .9404 & 395.26K
& 32.81 / .9535 & 430.17K
& 37.68 / .9630 & 351.22K \\
$\circ$ FINER
& 
& 
& 41.78 / .9775 & 309.20K
& 42.15 / .9803 & 305.60K
& 40.81 / .9876 & 325.74K
& 41.08 / .9796 & 312.68K \\
\SetCell[c=11]{l}Fixed-frequency, band-limited spectral INR Baselines \\
$\circ$ BACON
& \textbf{791K}
& \textbf{1000}
& 33.71 / .9488 & 390.47K
& 31.93 / .9194 & 429.31K
& 29.73 / .9177 & 462.32K
& 33.15 / .9414 & 390.39K \\
\SetCell[c=11]{l}Learnable-frequency spectral INR Baselines \\
$\circ$ FourierNet
& \SetCell[r=3]{c}\textbf{791K}
& \SetCell[r=3]{c}\textbf{1000}
& 45.67 / .9924 & 242.83K
& 41.77 / .9882 & 293.41K
& 39.31 / .9883 & 334.63K
& 42.64 / .9898 & 266.08K \\
$\circ$ GaborNet
& 
& 
& 50.31 / .9970 & 190.86K
& 51.38 / .9975 & 180.71K
& 49.51 / .9976 & 206.54K
& 49.54 / .9962 & 193.25K \\
$\circ$ iSIREN
& 
& 
& 51.80 / .9976 & 172.55K 
& 52.67 / .9981 & 167.44K 
& 51.43 / .9983 & 183.92K 
& 50.90 / .9971 & 173.83K \\
\SetCell[c=11]{l}Ours (Harmonic-Siren) \\
\SetCell[r=2]{halign=l, valign=m}\textbf{$\circ$ OURS}
& \SetCell[r=2]{c}\textbf{609K}
& \textbf{100}
& \textbf{58.16 / .9997} & \textbf{0.78K}
& \textbf{53.20 / .9989} & \textbf{2.51K}
& \textbf{46.28 / .9973} & \textbf{5.66K}
& \textbf{61.39 / .9998} & \textbf{0.62K} \\
& 
& \textbf{$\boldsymbol{\le}$ 1000}
& \SetCell[c=2]{c}\textbf{$\infty$ AT $322 \pm 117$ iters} & 
& \SetCell[c=2]{c}\textbf{$\infty$ AT $447 \pm 327$ iters} & 
& \SetCell[c=2]{c}\textbf{$\infty$ AT $537 \pm 432$ iters} & 
& \SetCell[c=2]{c}\textbf{$\infty$ AT $440 \pm 350$ iters} \\
\bottomrule
\end{tblr}
}
\label{tab:main_result_clean}
\vspace{-21pt}
\end{table*}

\paragraph{\textbf{Image Fitting.}}
Table~\ref{tab:main_result_clean} shows that finite sinusoidal refinement accelerates high-fidelity fitting under reduced parameter and optimization budget. With $609$K parameters and only $100$ iterations, \textit{ours} exceeds the $1{,}000$-iteration results of all learnable-frequency spectral baselines~\cite{Lindell_2022_CVPR,fathony2021multiplicative,NEURIPS2021_4ffbd5c8} on various datasets. This advantage is obtained despite using fewer parameters than the baselines. Fig.~\ref{fig:main_qual} shows the same early-optimization trend: compared with learnable-frequency baselines, \textit{ours} recovers sharper boundaries and finer structures with fewer residual artifacts, indicating that shared sinusoidal refinement improves the composition of learned harmonic components across recurrent steps.

\paragraph{\textbf{Super-resolution.}}
\tikzset{
  zoom/outer/.style={white, line width=0.30mm},
  zoom/inner/.style={white, line width=0.10mm},
  zoom/box/.style={
    opacity=1.0,
    height=1.35cm,
    width=1.35cm,
    connect spies,
    white
  },
}
\newcommand{\InterpZoomTriplet}[2]{%
  \begin{tikzpicture}[x=6cm, y=6cm,
    spy using outlines={every spy on node/.append style={zoom/inner}}]
    \node[anchor=south] (A) at (0,0) {%
      \includegraphics[trim=0 0 0 0, clip, height=1.5cm, keepaspectratio]{#1}%
    };

    \spy [zoom/box, magnification=10, size=.825cm] on ($(A)+(-0.000,+0.075)$)
      in node[zoom/outer, anchor=east, size=.825cm] at ($(A.north)+(0.02,-0.185)$);

    \spy [zoom/box, magnification=6.5, size=.6cm] on ($(A)+(+0.10,-0.015)$)
      in node[zoom/outer, anchor=east, size=.45cm] at ($(A.north)+(0.11,-0.225)$);

    \node[anchor=south, font=\footnotesize] at ($(A.north)+(0,-0.02)$) {#2};
  \end{tikzpicture}%
}
\begin{table*}[b!]
\vspace{-14pt}
  \centering
\caption{\textbf{Super-resolution with recurrent INRs.} We evaluate our framework for (a) image-specific INR fitting and (b) pretrained LIIF-based arbitrary-scale SR. For (b), we retain the EDSR-baseline encoder and replace only LIIF's coordinate MLP with our recurrent decoder under a matched $\sim$50K-parameter decoder budget.}
  \vspace{-20pt}
  \begin{subtable}[t]{0.358\textwidth}
    \centering
    \scriptsize
    \setlength{\tabcolsep}{2.0pt}
    \renewcommand{\arraystretch}{0.80}
    \caption{Per-image INR $\times2$.}
    \vspace{-10pt}
    \resizebox{\linewidth}{!}{%
    \begin{tblr}{
      colspec = {l c c c c},
      row{1} = {font=\bfseries},
      column{1-5} = {c},
      rowsep = 1.5pt,
      colsep = 2pt,
      row{4} = {bg=OursSoft},
      abovesep = .5pt,
      belowsep = .5pt,
    }
      \toprule
      Model & \#Params. & \#Iter. & PSNR$\uparrow$ & LPIPS$\downarrow$ \\
      \midrule
      RFF-F10 & 630.4K & 1000 & 25.37 & 0.4477 \\
      SIREN   & 611.6K & 1000 & \textbf{27.86} & 0.3068 \\
      \textbf{Ours} & 609K & $595\!\pm\!285$ & 26.55 & \textbf{0.2272} \\
      \bottomrule
    \end{tblr}}
\vspace{-.5pt}
    \resizebox{\linewidth}{!}{%
      \InterpZoomTriplet{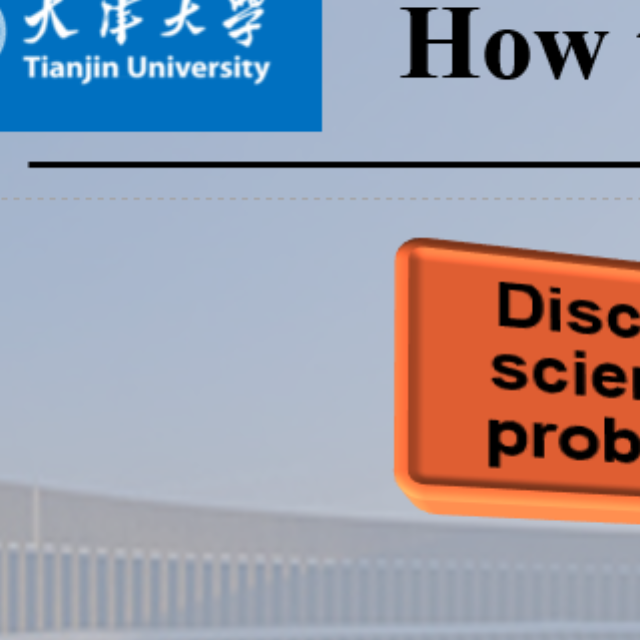}{\tiny Bilinear $\times2$}%
      \hspace{-1.5mm}%
      \InterpZoomTriplet{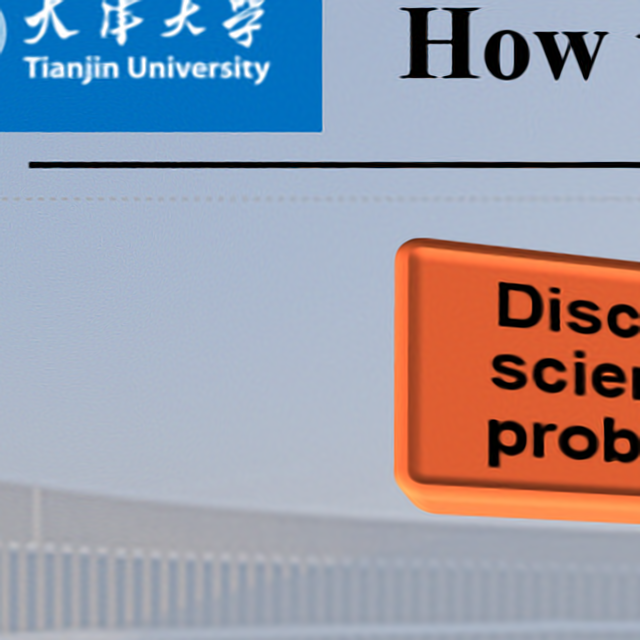}{\tiny SIREN $\times2$}%
      \hspace{-1.5mm}%
      \InterpZoomTriplet{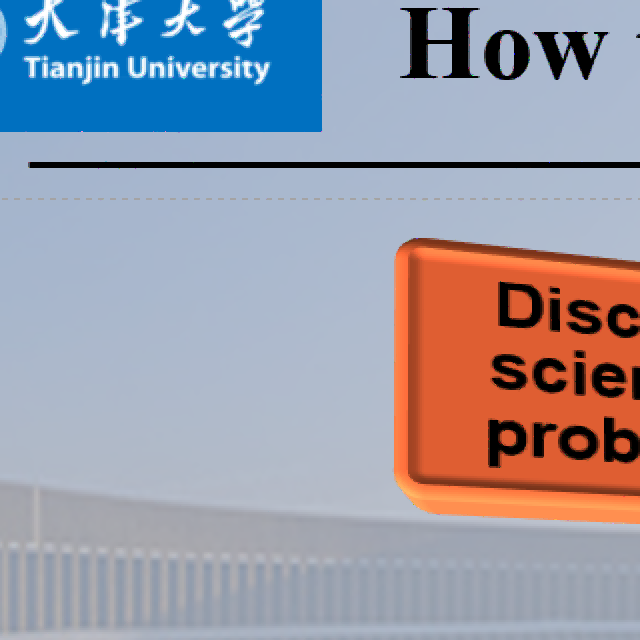}{\tiny Ours $\times2$}%
    }
    \label{fig:sr-perimage}
  \end{subtable}%
  \hfill
  \begin{subtable}[t]{0.622\textwidth}
    \centering
    \scriptsize
    \caption{Pretrained LIIF SR (arbitrary-scale).}
    \vspace{-10pt}
    \resizebox{\linewidth}{!}{%
    \begin{tblr}{
      colspec = {l c c c c | c c c},
      row{1-2} = {font=\bfseries},
      abovesep = .5pt,
      belowsep = .5pt,
      row{3} = {bg=OursSoft},
    }
      \toprule
      \SetCell[r=2]{c}{Dataset}
      & \SetCell[r=2]{c}{Scale}
      & \SetCell[c=3]{c}{LIIF MLP ($\approx$50K)}
      & &
      & \SetCell[c=3]{c}{\textbf{\textit{Ours}} ($\approx$50K)}
      & &
      \\
      &
      & PSNR$\uparrow$ & SSIM$\uparrow$ & LPIPS$\downarrow$
      & PSNR$\uparrow$ & SSIM$\uparrow$ & LPIPS$\downarrow$
      \\
      \midrule
      \SetCell[c=8]{l}{\textit{Benchmark (EDSR-baseline encoder)~\cite{Lim_2017_CVPR_Workshops}}} \\
      \SetCell[r=3]{c}{Set5}
      & $\times2$ & 37.673 & 0.9485 & 0.0625 & 37.721 & 0.9489 & 0.0574 \\
      & $\times3$ & 34.036 & 0.9104 & 0.1292 & 34.103 & 0.9111 & 0.1275 \\
      & $\times4$ & 31.820 & 0.8711 & 0.1797 & 31.850 & 0.8718 & 0.1867 \\
      \midrule
      \SetCell[r=3]{c}{Set14}
      & $\times2$ & 33.325 & 0.9005 & 0.1022 & 33.401 & 0.9016 & 0.0941 \\
      & $\times3$ & 30.116 & 0.8243 & 0.2121 & 30.126 & 0.8249 & 0.2103 \\
      & $\times4$ & 28.377 & 0.7582 & 0.2912 & 28.389 & 0.7598 & 0.3010 \\
      \midrule
      \SetCell[r=3]{c}{B100}
      & $\times2$ & 31.968 & 0.8979 & 0.1624 & 32.031 & 0.8996 & 0.1524 \\
      & $\times3$ & 28.926 & 0.8046 & 0.2966 & 28.976 & 0.8060 & 0.2969 \\
      & $\times4$ & 27.427 & 0.7294 & 0.3856 & 27.450 & 0.7308 & 0.3981 \\
      \bottomrule
    \end{tblr}}
    \label{fig:sr-amortized}
  \end{subtable}
  \label{fig:sr-comparison}
\end{table*}
We evaluate unseen-coordinate recovery in complementary image-specific and pretrained settings. For per-image $\times2$ fitting (Table~\ref{fig:sr-perimage}), \textit{ours} preserves a sharper vertical boundary on screen content than SIREN~\cite{NEURIPS2020_53c04118} and bilinear interpolation, and attains the lowest Kodak24 LPIPS~\cite{Zhang_2018_CVPR} with fewer optimization iterations on average.
For pretrained LIIF-based arbitrary-scale SR~\cite{Chen_2021_CVPR} (Table~\ref{fig:sr-amortized}), we retain the EDSR-baseline encoder~\cite{Lim_2017_CVPR_Workshops} and replace only the coordinate MLP under a matched $\sim$50K-parameter decoder budget. Across Set5, Set14, and B100 at $\times2$--$\times4$, our decoder consistently improves PSNR and SSIM, while perceptual gains are most pronounced at $\times2$.

\begin{table*}[t!]
\caption{\textbf{NVS quality on four scenes.}
We compare NeRF and \textit{Ours} under the same training budget; entries report PSNR, SSIM, and LPIPS for the rendered RGB views. Consistent gains across scenes indicate improved geometry-aware reconstruction.}
\label{tab:nerf_rec_bin}
\centering
\scriptsize
\vspace{-10pt}
\captionsetup[subtable]{
  position=top,
  justification=centering,
  singlelinecheck=false,
  skip=2pt
}

\begin{subtable}[t]{0.49\linewidth}
\centering
\begin{tblr}{
  width = \linewidth,
  colspec = {X[1.25,l] X[1,c] X[1,c] X[1,c]},
  row{1} = {font=\bfseries, 
  abovesep = 0pt,
  belowsep = 0pt,},
  rowsep = 2pt,
  colsep = 2pt,
  abovesep = 1pt,
  belowsep = 1pt,
}
\toprule
\textbf{NeRF} & PSNR $\uparrow$ & SSIM $\uparrow$ & LPIPS $\downarrow$ \\
\midrule
Flower   & 23.71 & 0.6316 & 0.4145 \\
Orchids  & 17.85 & 0.4198 & 0.5172 \\
Fern     & 21.06 & 0.5640 & 0.4943 \\
Fortress & 26.10 & 0.5987 & 0.4423 \\
\bottomrule
\end{tblr}
\end{subtable}
\hfill
\begin{subtable}[t]{0.49\linewidth}
\centering
\begin{tblr}{
  width = \linewidth,
  colspec = {X[1.25,l] X[1,c] X[1,c] X[1,c]},
  row{1} = {font=\bfseries, 
  abovesep = 0pt,
  belowsep = 0pt,},
  rowsep = 2pt,
  colsep = 2pt,
  abovesep = 1pt,
  belowsep = 1pt,
}
\toprule
\textbf{+ Ours} & PSNR $\uparrow$ & SSIM $\uparrow$ & LPIPS $\downarrow$ \\
\midrule
Flower   & 24.54 & 0.7138 & 0.2934 \\
Orchids  & 18.79 & 0.5313 & 0.3797 \\
Fern     & 21.60 & 0.6500 & 0.3704 \\
Fortress & 26.36 & 0.7189 & 0.3206 \\
\bottomrule
\end{tblr}
\end{subtable}

\vspace{-15pt}
\end{table*}
\newcommand{\NVSImgH}{0.60in}
\newcommand{\NVSPairGap}{-1.45mm}
\newcommand{\NVSColGap}{-2.45mm}
\newcommand{\NVSRowLeftPad}{-1.25mm}
\newcommand{\NVSRowLabelGap}{-0.5mm}
\newcommand{\NVSRowVSpace}{-5pt}
\newcommand{\NVSTitleShift}{-1mm}

\newcommand{\NVSRowLabel}[2]{%
  \hspace{\NVSRowLeftPad}%
  \raisebox{#1}{\rotatebox{90}{\scriptsize #2}}%
  \hspace{\NVSRowLabelGap}%
}

\newcommand{\NVSPairWithTitle}[3]{%
\begin{tikzpicture}[x=1cm,y=1cm]
  \node[anchor=south] (pair) at (0,0) {%
    \includegraphics[height=\NVSImgH,keepaspectratio]{#2}%
    \hspace{\NVSPairGap}%
    \includegraphics[height=\NVSImgH,keepaspectratio]{#3}%
  };
  \node[anchor=south, yshift=\NVSTitleShift] at (pair.north) {\scriptsize #1};
\end{tikzpicture}%
}

\newcommand{\NVSPair}[2]{%
\begin{tikzpicture}[x=1cm,y=1cm]
  \node[anchor=south] (pair) at (0,0) {%
    \includegraphics[height=\NVSImgH,keepaspectratio]{#1}%
    \hspace{\NVSPairGap}%
    \includegraphics[height=\NVSImgH,keepaspectratio]{#2}%
  };
\end{tikzpicture}%
}

\begin{figure*}[t]
\footnotesize
\centering
\vspace{3pt}
\NVSRowLabel{0.22in}{NeRF}%
\NVSPairWithTitle{Orchids}%
  {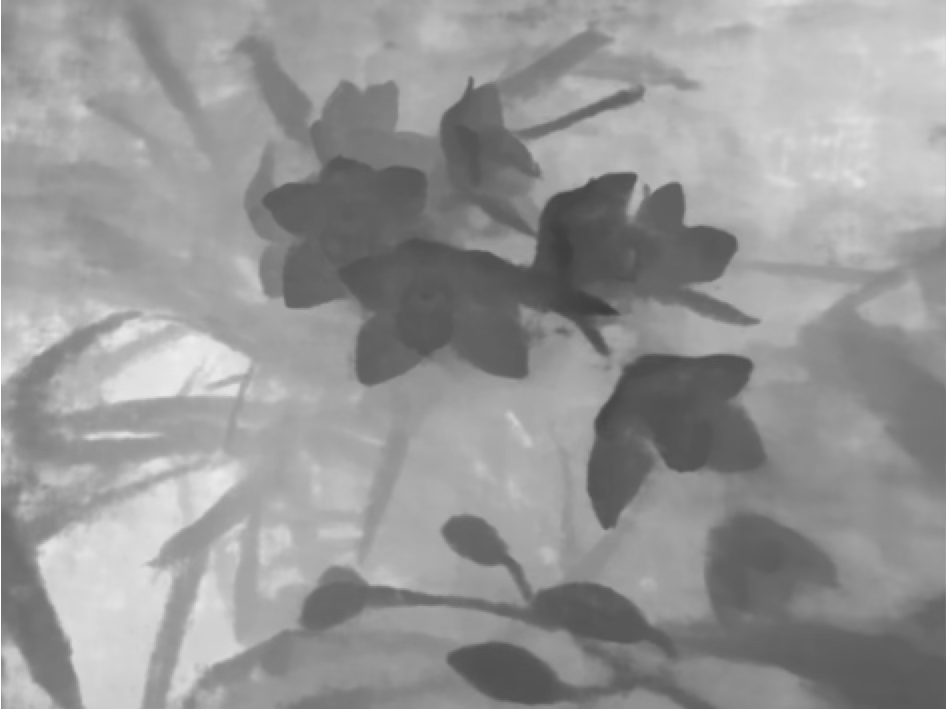}%
  {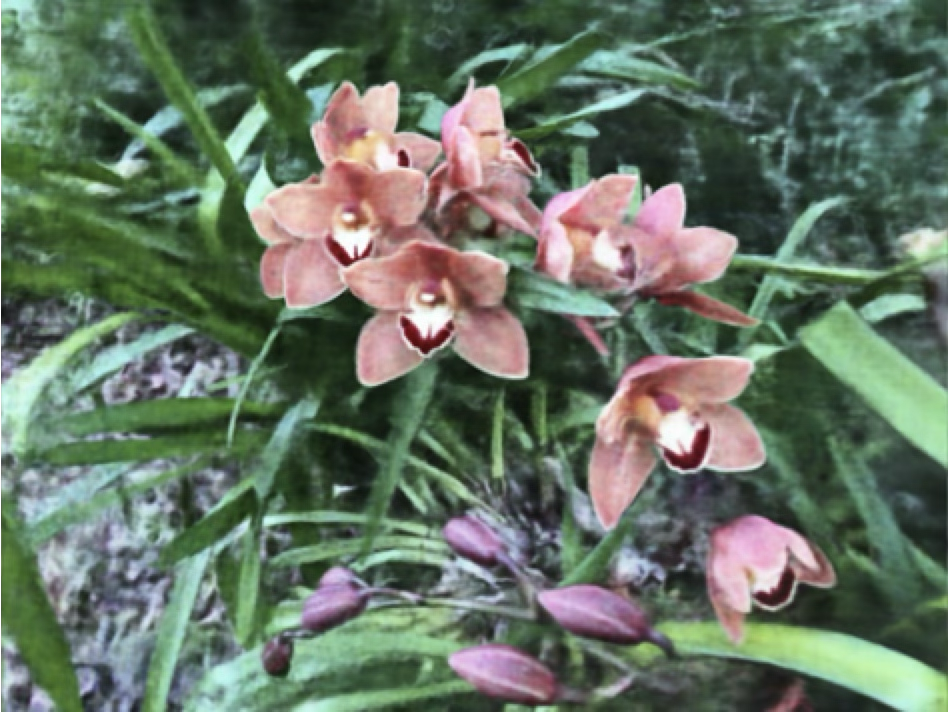}%
\hspace{\NVSColGap}%
\NVSPairWithTitle{Flower}%
  {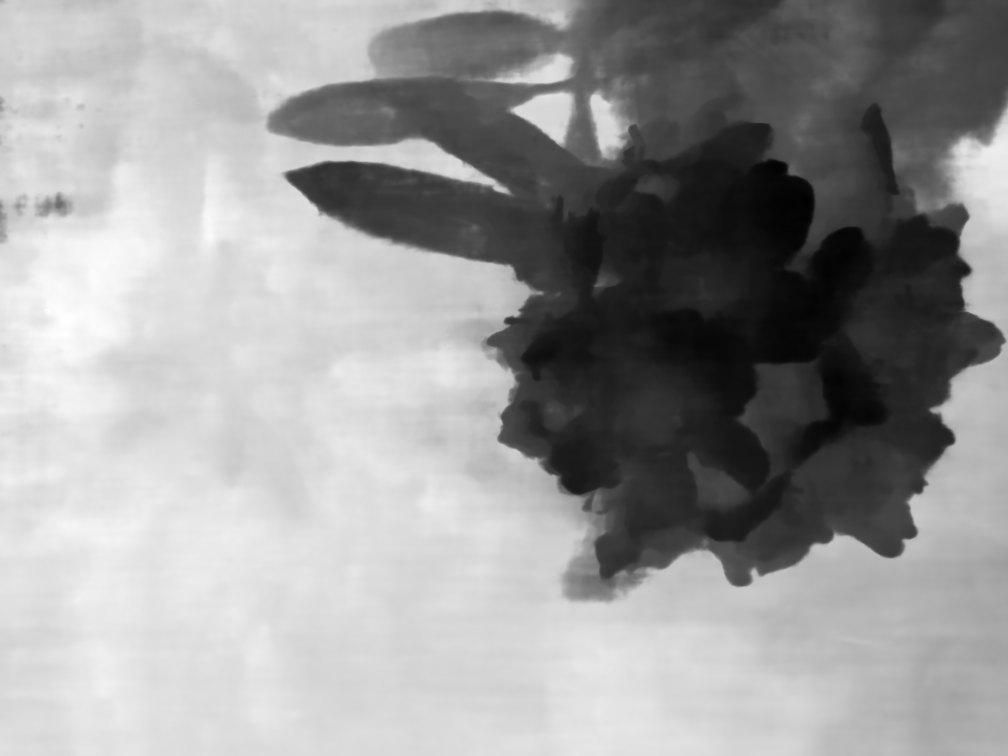}%
  {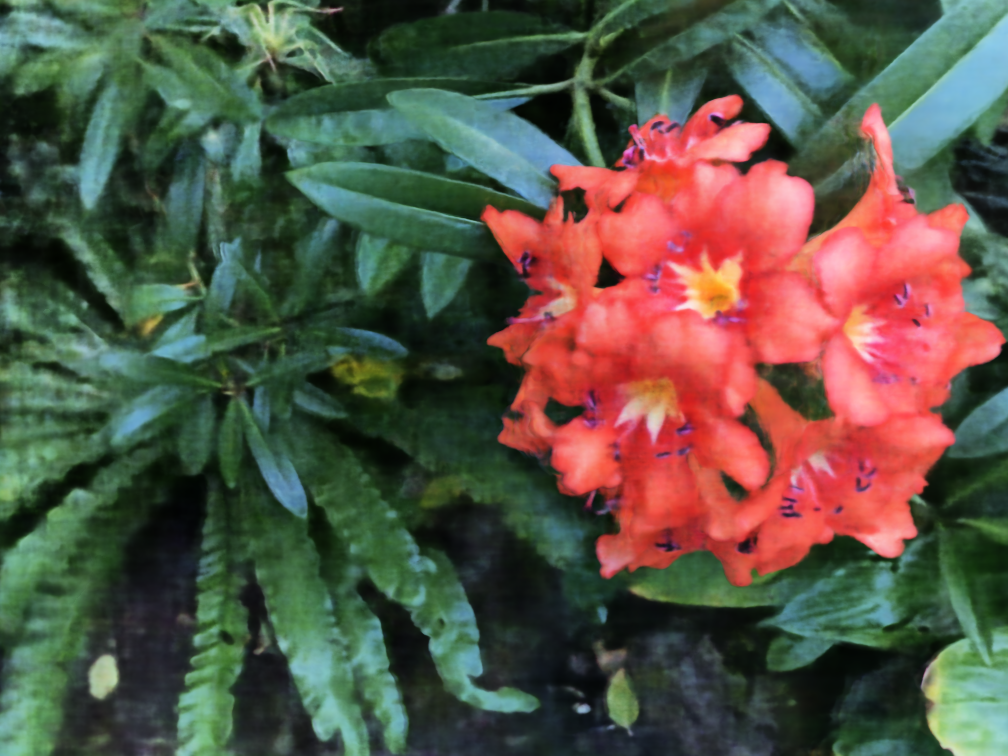}%
\hspace{\NVSColGap}%
\NVSPairWithTitle{Fern}%
  {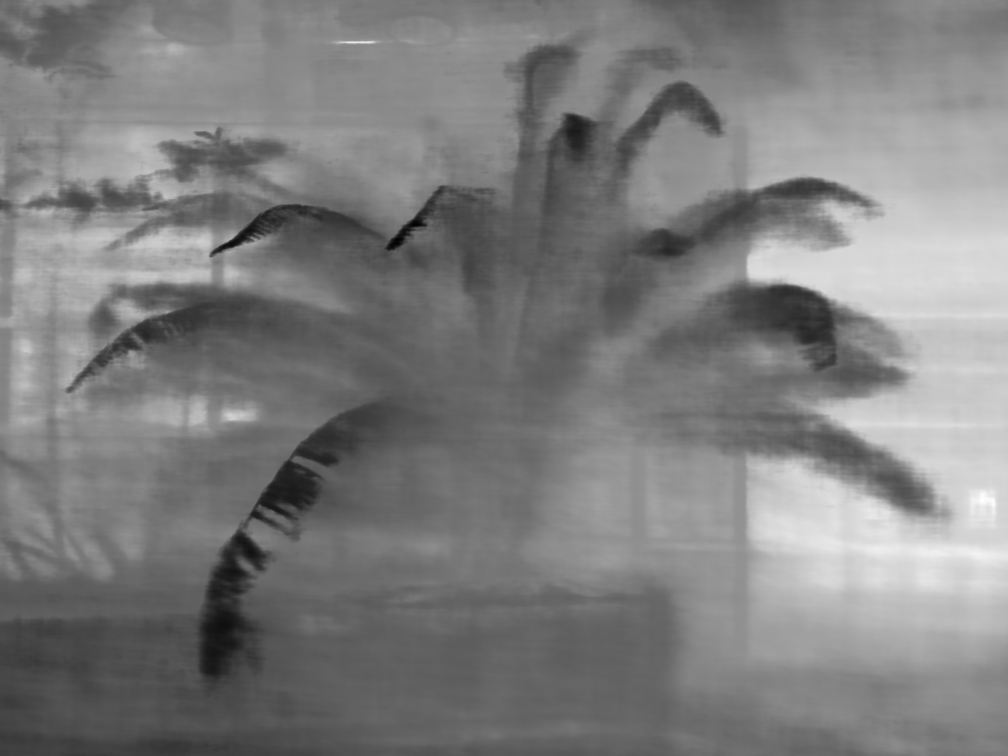}%
  {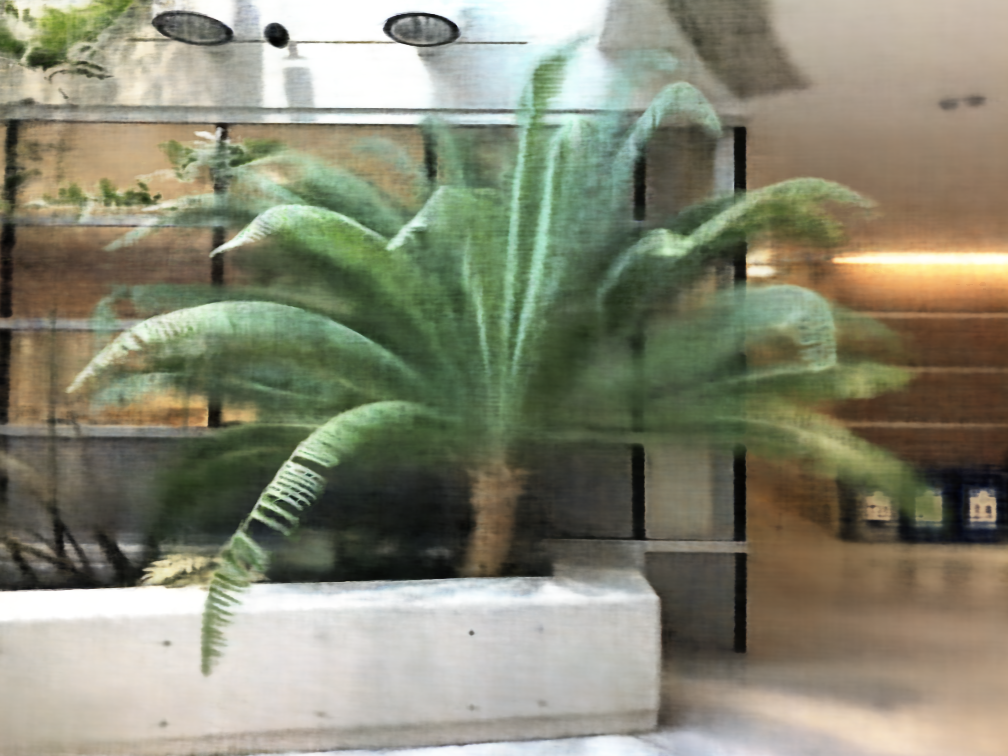}%

\vspace{\NVSRowVSpace}

\NVSRowLabel{0.16in}{\textit{Ours}}%
\NVSPair%
  {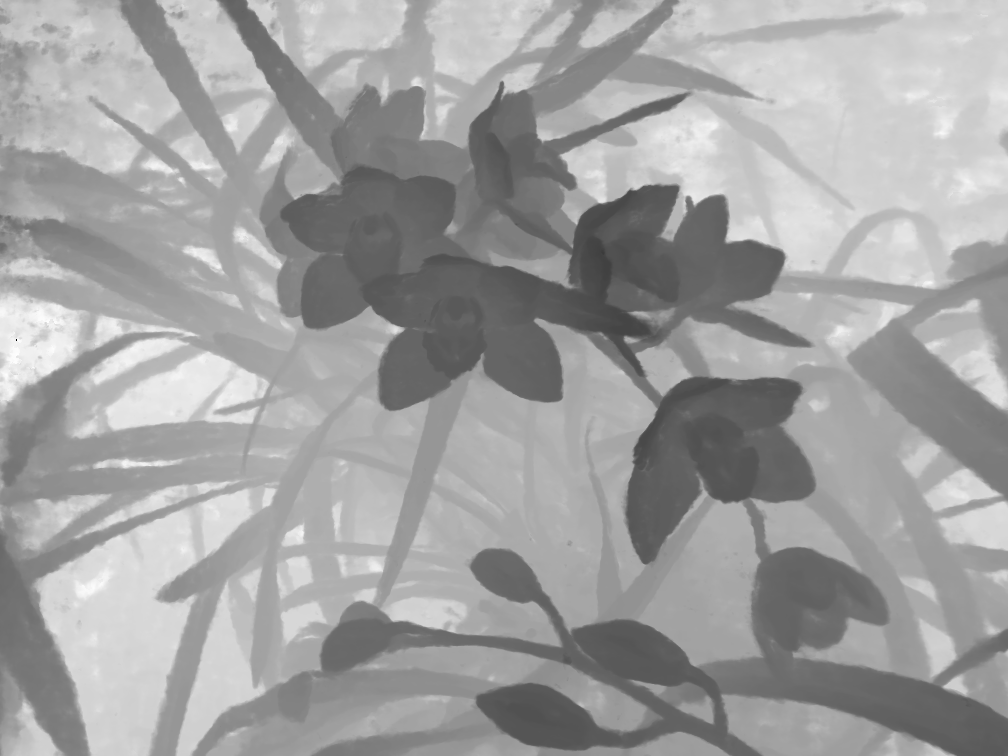}%
  {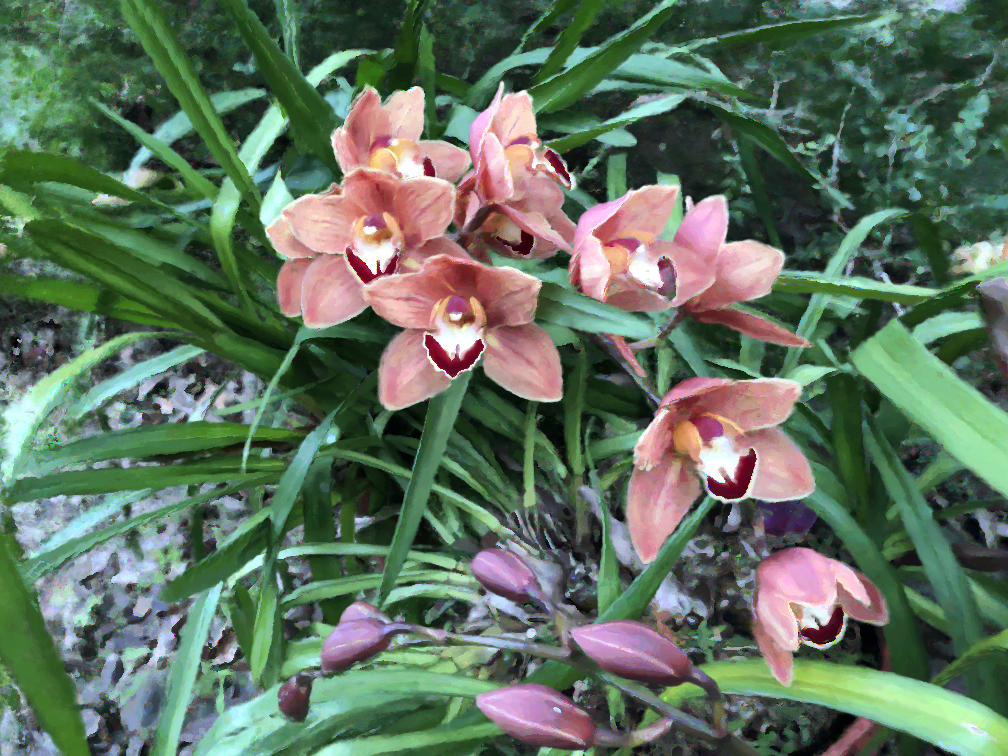}%
\hspace{\NVSColGap}%
\NVSPair%
  {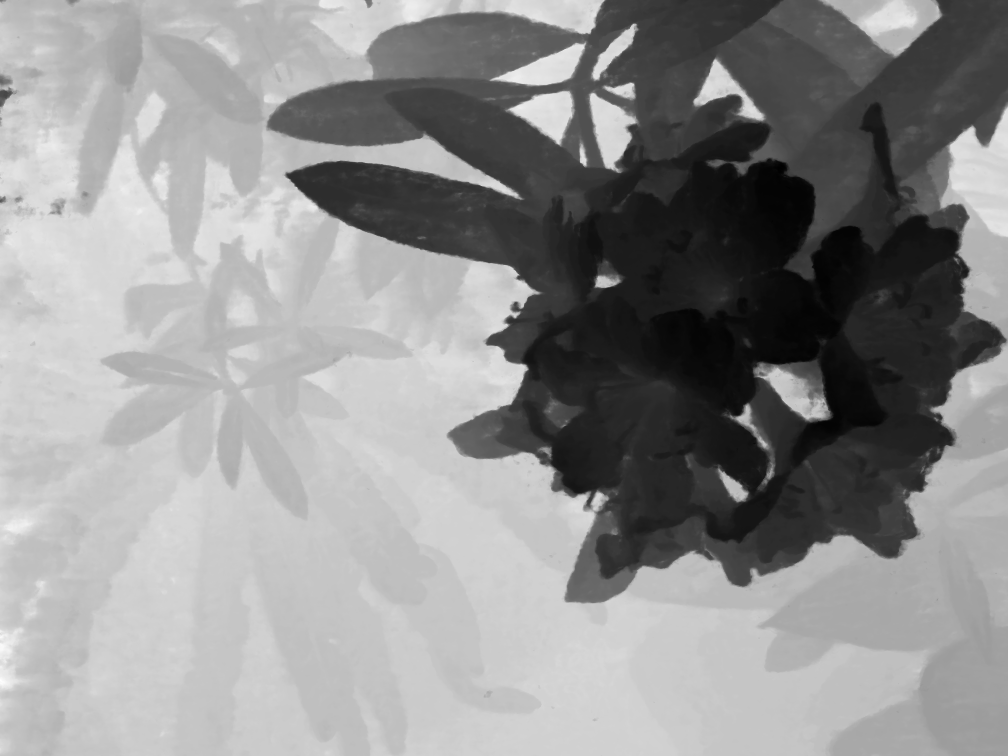}%
  {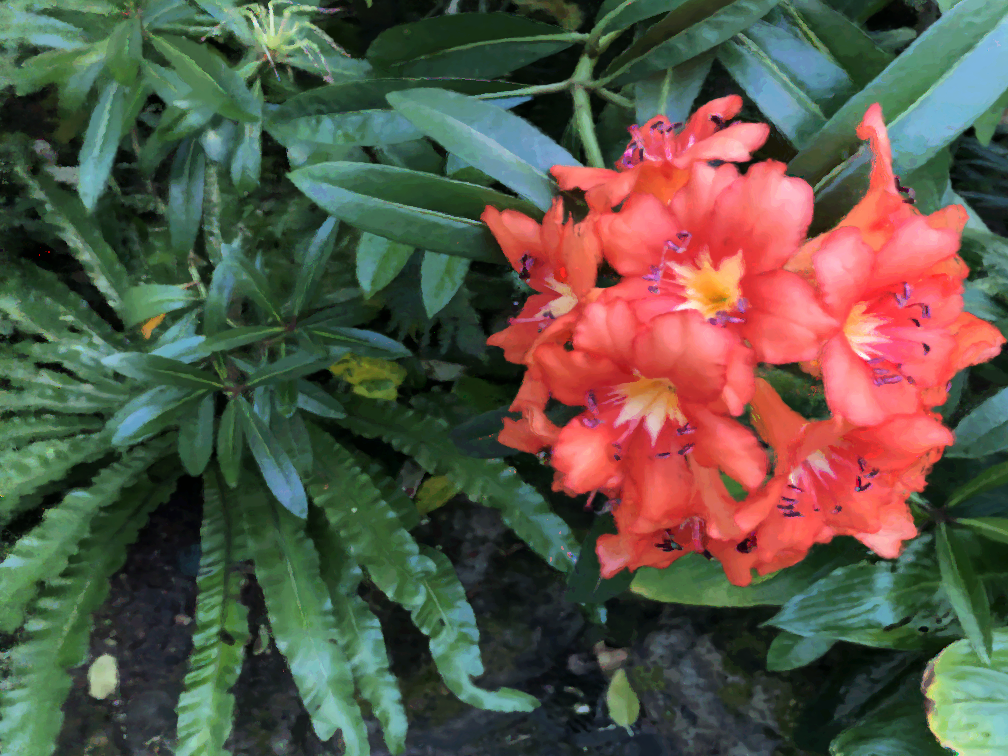}%
\hspace{\NVSColGap}%
\NVSPair%
  {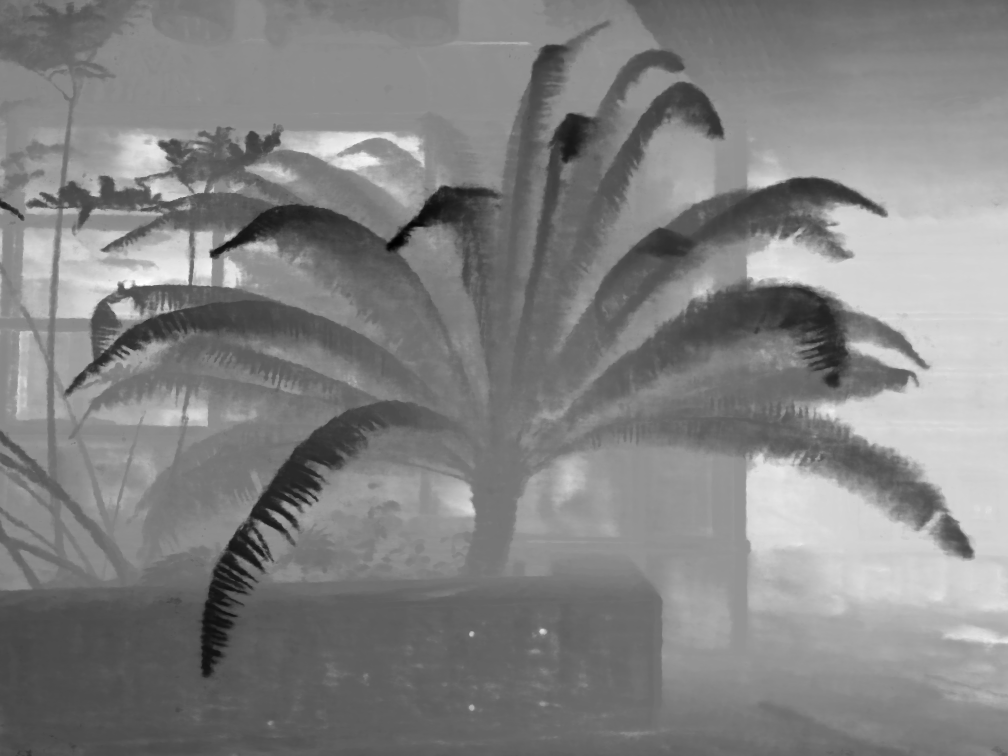}%
  {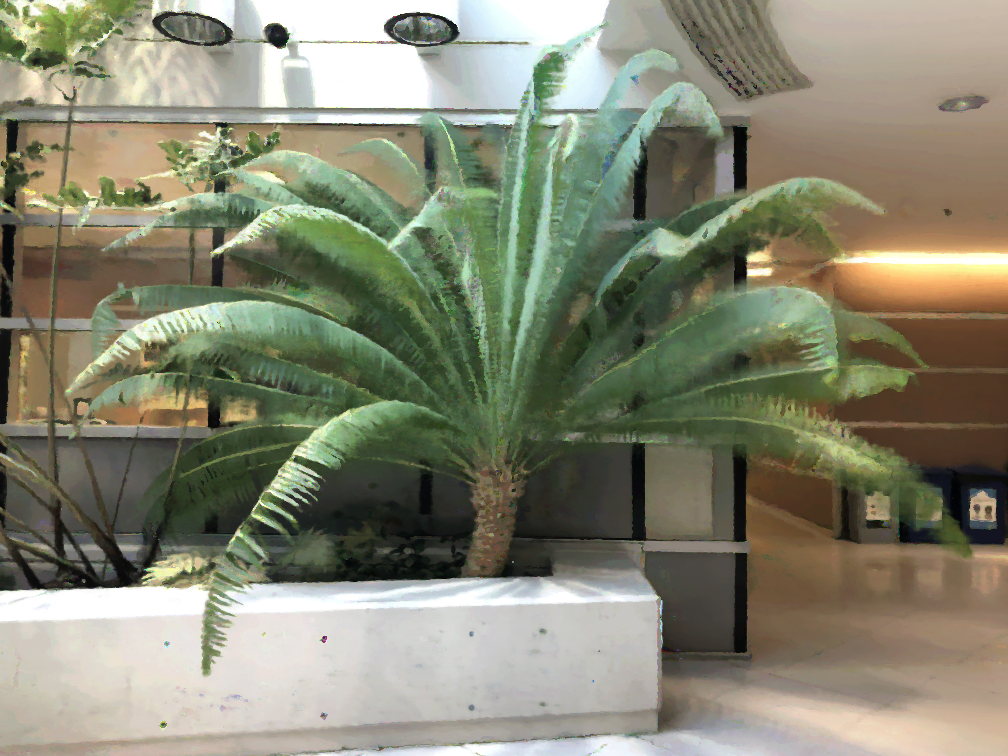}%
  \vspace{-10pt}
\caption{\textbf{Qualitative novel view synthesis results (depth + rendering).}
We compare NeRF and \textit{Ours} on three scenes, showing predicted depth (grayscale) alongside the corresponding rendered RGB views. Under the same model budget (600K parameters) and training protocol, \textit{Ours} yields cleaner, more coherent depth with sharper object boundaries, which translates to crisper geometry-aware renderings.}
\label{fig:nerf_qualitative_results}
\vspace{-16.5pt}
\end{figure*}

\subsection{Extension to 3D Representation}
\paragraph{\textbf{Neural Radiance Field.}}
We evaluate \textit{Ours} in a neural radiance field setting on the forward-facing LLFF dataset~\cite{llff}. We replace the standard NeRF RGB decoder with our recurrent sinusoidal decoder while keeping the rest of the rendering pipeline unchanged and maintaining a comparable parameter budget. As shown in Table~\ref{tab:nerf_rec_bin}, \textit{Ours} consistently improves PSNR, SSIM~\cite{1284395}, and LPIPS~\cite{Zhang_2018_CVPR} over the baseline NeRF~\cite{10.1007/978-3-030-58452-8_24}. The lower LPIPS is consistent with better preservation of fine appearance details and local sharpness. This trend is also visually reflected in Fig.~\ref{fig:nerf_qualitative_results}, where \textit{Ours} produces visually sharper renderings and more coherent depth predictions, suggesting that the proposed decoder benefits both view synthesis quality and scene structure recovery.

\paragraph{\textbf{Signed Distance Function.}}
\tikzset{
    largewindow_w/.style={orange, line width=0.30mm},
    smallwindow_w/.style={orange, line width=0.10mm},
    closeup_w_3/.style={
        opacity=1.0,
        height=1.35cm,
        width=1.35cm,
        connect spies,
        orange
    },
}

\newcommand{\ModelImageSDF}[2]{%
    \begin{tikzpicture}[x=5cm, y=5cm, spy using outlines={every spy on node/.append style={smallwindow_w}}]
        \node[anchor=south] (FigA) at (0,0) {\includegraphics[trim=0 0 0 0, clip, height=1.4cm, keepaspectratio]{#1}};
        \spy [closeup_w_3, magnification=2.4, size=1.0cm] on ($(FigA)+(-0.17,0)$) 
            in node[largewindow_w, anchor=east, size=1.1cm] at ($(FigA.north)+(0.23,-0.15)$);
    \end{tikzpicture}
}
\newcommand{\ModelImageSDFSec}[2]{%
    \begin{tikzpicture}[x=5cm, y=5cm, spy using outlines={every spy on node/.append style={smallwindow_w}}]
        \node[anchor=south] (FigA) at (0,0) {\includegraphics[trim=0 0 0 0, clip, height=1.4cm, keepaspectratio]{#1}};
        \spy [closeup_w_3, magnification=3, size=1.0cm] on ($(FigA)+(0.145,-0.065)$)
            in node[largewindow_w, anchor=east, size=.80cm] at ($(FigA.north)+(0.05,-0.21)$);
    \end{tikzpicture}
}

\begin{wrapfigure}{r}{0.47\linewidth}
    \vspace{-25pt}
    \footnotesize
    \centering
    \hspace{-2.0mm}
    \raisebox{0.2in}{\rotatebox{90}{\scriptsize\textit{Ours}}}%
    \hspace{-1.5mm}
    \ModelImageSDF{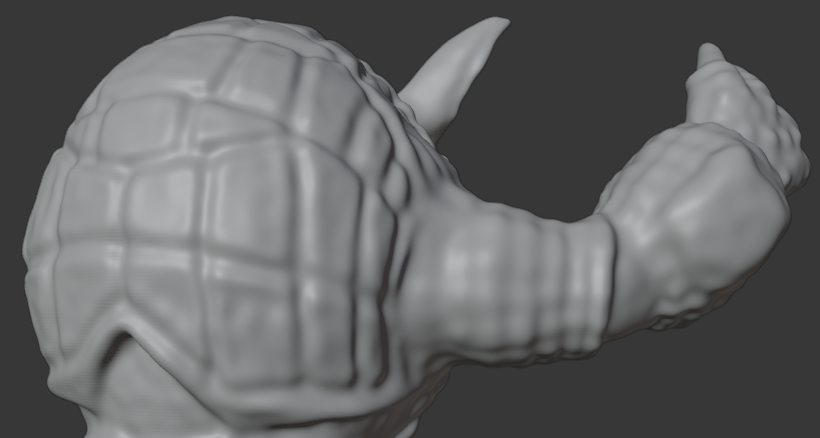}{20.28}
    \hspace{-3.8mm}
    \ModelImageSDFSec{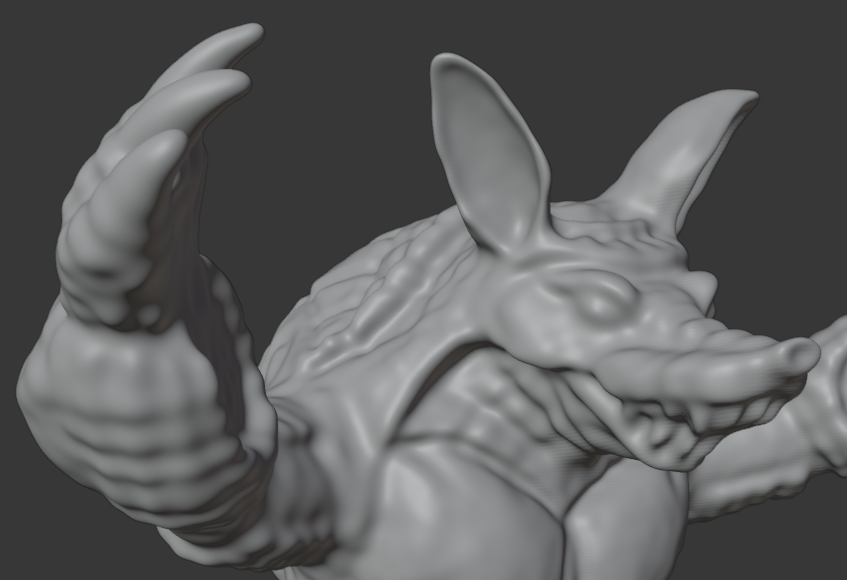}{27.15}
    
    \vspace{-6pt}
    
    \hspace{-1.8mm}
    \raisebox{0.165in}{\rotatebox{90}{\scriptsize$\mathrm{SIREN}$}}%
    \hspace{-1.8mm}
    \ModelImageSDF{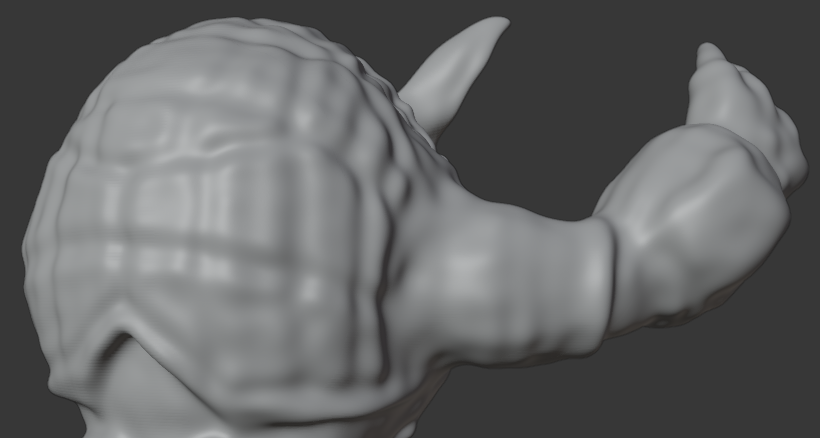}{7.08}
    \hspace{-3.8mm}
    \ModelImageSDFSec{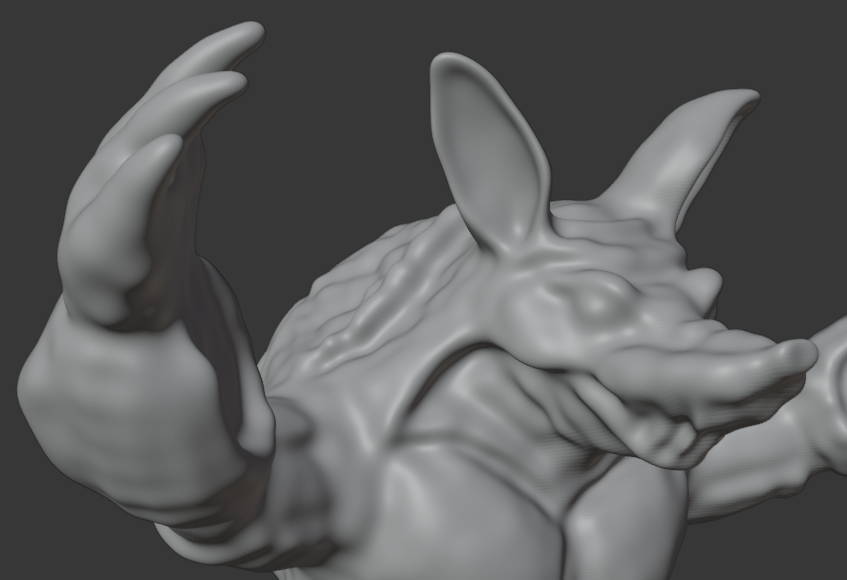}{10.52}
    \caption{\textbf{Ours on SDF.} Under the same training budget, Ours reconstructs finer surface structure than SIREN~\cite{NEURIPS2020_53c04118}.}
    \vspace{-30pt}
    \label{fig:gamsdf}
\end{wrapfigure}
We evaluate \textit{Ours} on signed distance function (SDF) reconstruction using the Stanford Armadillo dataset~\cite{10.1145/237170.237270}, following the setup of \textit{Sitzmann} \etal~\cite{NEURIPS2020_53c04118}. We compare against SIREN under the same training budget of 2K epochs. Our SDF model uses a single hidden layer with 500 units and three recurrent steps. As shown in Fig.~\ref{fig:gamsdf}, \textit{Ours} recovers finer surface structure and preserves sharper geometric details than SIREN, indicating that the proposed recurrent sinusoidal decoder transfers effectively to continuous 3D shape representation.

\section{Discussion and Limitations}
\label{sec:discussion}
\begin{table}[t]
\centering
\footnotesize
\setlength{\tabcolsep}{3.0pt}
\renewcommand{\arraystretch}{1.12}

\begin{minipage}[t]{0.48\linewidth}
\centering
\captionof{table}{\textbf{Effect of Gray coding on 2D image $\times2$ super-resolution.} Replacing natural binary coding (NBC) with Gray coding (GC) improves both reconstruction at the training resolution and generalization to unseen coordinates for the same architecture and parameter budget.}
\label{tab:gray_2d}

\resizebox{\linewidth}{!}{%
\begin{tblr}{
  colspec = {l c c c c},
  row{1}  = {font=\bfseries},
  column{2-5} = {c},
  rowsep  = 2pt,
  colsep  = 4pt,
  row{3} = {bg=LightBeige},
  row{5} = {bg=OursSoft},
}
\toprule
Architecture & Coding & \#Params. & Fit PSNR$\uparrow$ & SR PSNR $\times2$$\uparrow$ \\
\midrule
\SetCell[r=2]{l}{RFF-F10}      & NBC    & \SetCell[r=2]{c}{630.4K} & 22.35 & 21.08 \\
             & GC     &                          & 26.30 & 23.98 \\
\midrule
\SetCell[r=2]{l}{\textbf{Ours}} & NBC   & \SetCell[r=2]{c}{609K}   & $\infty$ & 23.45 \\
& GC    &                          & $\infty$ & \textbf{26.55} \\
\bottomrule
\end{tblr}}
\end{minipage}
\hfill
\begin{minipage}[t]{0.48\linewidth}
\centering
\captionof{table}{\textbf{Effect of Gray coding on NeRF rendering.}
Results on the Flower scene from LLFF. Replacing natural binary coding (NBC) with Gray coding (GC) improves PSNR and SSIM and lowers LPIPS for both the standard NeRF decoder and \textbf{Ours}.}
\label{tab:gray_nerf}

\resizebox{\linewidth}{!}{%
\begin{tblr}{
  colspec = {l c c c c c},
  row{1}  = {font=\bfseries},
  column{2-6} = {c},
  rowsep  = 2pt,
  colsep  = 4pt,
  row{3} = {bg=LightBeige},
  row{5} = {bg=OursSoft},
}
\toprule
Architecture & Coding & \#Params. & PSNR$\uparrow$ & SSIM$\uparrow$ & LPIPS$\downarrow$ \\
\midrule
\SetCell[r=2]{l}{Baseline NeRF} & NBC   & \SetCell[r=2]{c}{602.9K} & 20.79 & 0.5079 & 0.4465 \\
              & GC    &                          & 23.05 & 0.6612 & 0.3437 \\
\midrule
\SetCell[r=2]{l}{\textbf{Ours}} & NBC   & \SetCell[r=2]{c}{600K}   & 21.14 & 0.5263 & 0.4370 \\
 & GC    &                          & \textbf{23.80} & \textbf{0.7010} & \textbf{0.3030} \\
\bottomrule
\end{tblr}}
\end{minipage}

\vspace{-6pt}
\end{table}
\subsection{Code-space continuity via Gray coding}
\label{sec:gray}
Binary-domain supervision can represent quantized signals exactly, but its effectiveness depends critically on how quantization levels are mapped to codewords. Let $\Gamma:\{0,\ldots,2^n-1\}\to\{0,1\}^n$ be an $n$-bit encoder, and let $\tilde{\Gamma}(q)=2\Gamma(q)-1\in\{-1,+1\}^n$ denote its bipolar form. If two codewords differ in $h$ bits, then their bipolar cosine similarity is
\begin{equation}
\frac{\tilde{\Gamma}(q)^\top \tilde{\Gamma}(q')}{\|\tilde{\Gamma}(q)\|_2\,\|\tilde{\Gamma}(q')\|_2}
=
1-\frac{2h}{n}.
\label{eq:gray_cosine_hamming}
\end{equation}
Thus, in bipolar space, the Hamming distance directly determines how abruptly adjacent targets change under cosine supervision. To quantify this effect, we define the local code-space sensitivity for one quantization increment $\delta$ as
\begin{equation}
L_{\Gamma}^{\rm local}
=
\max_{q}
\frac{\mathcal{H}\!\left(\Gamma(q),\Gamma(q+1)\right)}{\delta},
\label{eq:local_code_sensitivity}
\end{equation}
where $\mathcal{H}$ denotes the Hamming distance and, for normalized intensities, $\delta=2^{-n}$ corresponds to one quantization step. Under natural binary coding (NBC), adjacent levels can differ in as many as $n$ bits, yielding
\begin{equation}
L_{\rm NBC}^{\rm local}=\frac{n}{\delta}.
\label{eq:local_nbc}
\end{equation}
In contrast, Gray coding~\cite{gray1953pulse} flips exactly one bit between adjacent levels by construction, so
\begin{equation}
L_{\rm GC}^{\rm local}=\frac{1}{\delta}.
\label{eq:local_gc}
\end{equation}
Therefore, Gray coding reduces the worst-case local sensitivity by a factor of $n$. 

We validate this effect in two settings. On Kodak image fitting and subsequent ($\times$ 2) super-resolution (Table~\ref{tab:gray_2d}), replacing NBC with Gray coding improves generalization to unseen coordinates for both a Fourier-feature baseline and Ours. On the Flower scene of LLFF NeRF rendering (Table~\ref{tab:gray_nerf}), Gray coding consistently improves PSNR and SSIM while reducing LPIPS, with the largest gains observed for Ours.

\begin{figure*}[t]
  \centering
  \begin{subfigure}{0.6\linewidth}
    \centering
    \includegraphics[width=1.\columnwidth]{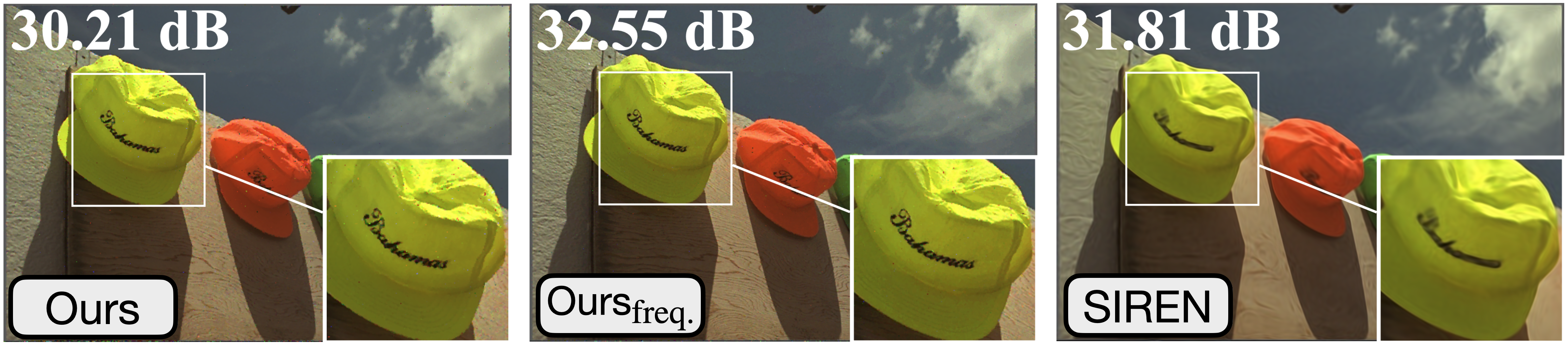}
    \caption{\textbf{Qualitative reconstruction (375K params, 1k steps).} The bit-plane reweighted \textit{Ours} attains higher PSNR than the unweighted variant and SIREN while better preserving fine local structures. In particular, the crops reveal sharper boundary transitions and more faithful recovery of the small printed markings on the yellow cap.}
    \label{fig:limitation1_qual}
  \end{subfigure}
  \hfill
  \begin{subfigure}{0.38\linewidth}
    \centering
    \includegraphics[width=.9\columnwidth]{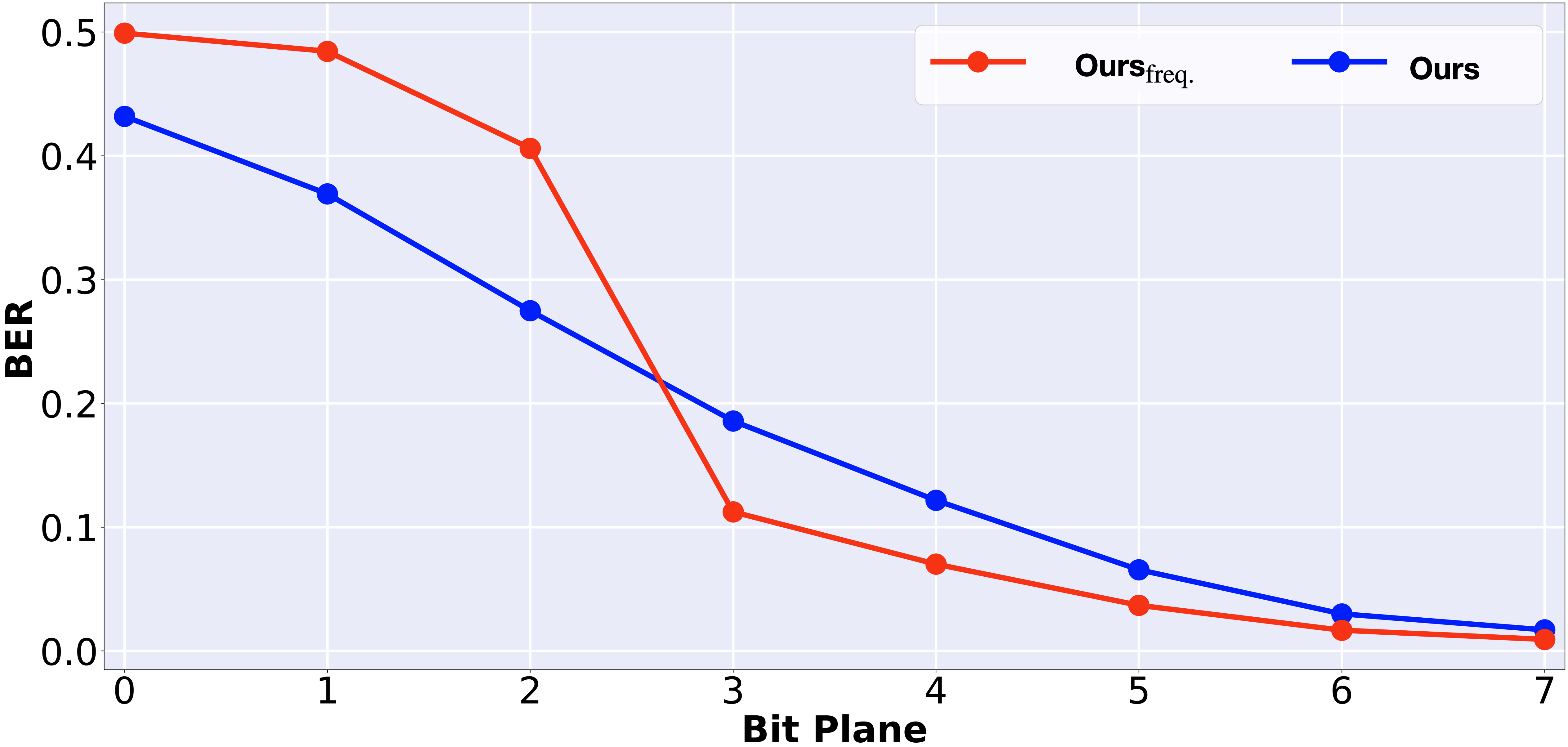}
    \caption{\textbf{Bit Error Rate (BER) profile.} Reweighting shifts BER toward less significant bits, reducing errors on higher-significance planes and improving overall fidelity.}
  \end{subfigure}
  \vspace{-5pt}
  \caption{\textbf{Bit-plane reweighting improves fidelity under a fixed parameter budget.}
(a) Reconstruction of a $768\times512$ Kodak image using 375K trainable parameters and 1,000 optimization steps. The reweighted \textit{Ours} achieves higher PSNR than both the unweighted model and SIREN, with sharper local details. (b) Bit-error rate (BER) across bit planes: reweighting prioritizes high-significance bits and shifts residual errors to less significant planes, improving image quality without increasing parameters.}
  \vspace{-13pt}
  \label{fig:supp_gamw}
\end{figure*}

\subsection{Decoded-Distortion-Aware Bit-Plane Reweighting}
The Gray-code analysis in Sec.~\ref{sec:gray} also distinguishes residual code errors by their effect after decoding. When the model or optimization budget is insufficient for exact reconstruction, residual errors need not be equally costly in the reconstructed signal. In particular, errors in lower-significance Gray planes induce bounded changes only in lower-order decoded binary digits, whereas errors in higher-significance planes can produce substantially larger intensity deviations. We therefore redistribute optimization emphasis toward the more consequential code planes in the non-exact regime. Concretely, we use a weighted loss{
\setlength{\abovedisplayskip}{3pt}
\setlength{\belowdisplayskip}{3pt}
\begin{equation}
\mathcal{L}_{\mathrm{w}}
=
\sum_{i=0}^{B-1} w_i\,\mathcal{L}^{(i)},
\label{eq:weighted_plane_loss}
\end{equation}}where $\mathcal{L}^{(i)}$ is the loss for the $i$-th code plane. In our implementation, with $i=0$ denoting the least significant plane, we set{
\setlength{\abovedisplayskip}{3pt}
\setlength{\belowdisplayskip}{3pt}
\begin{equation}
w_i = 2^{-(8-i)}, \qquad i\in\{0,1,2\},
\qquad
w_i = 1,\quad \text{otherwise}.
\label{eq:weighted_plane_coeff}
\end{equation}}This reweighting effectively reallocates the available budget during optimization. As shown in Fig.~\ref{fig:supp_gamw}, it improves reconstruction quality under the same parameter budget.

\vspace{-10pt}

\subsection{Iterative refinement and wall-clock efficiency}
Because recurrent unrolling increases the computation performed at each optimization step, we evaluate the wall-clock efficiency of \textit{Ours} on Kodak24. As shown in Table~\ref{tab:supp_timeconsumption}, \textit{Ours} already reaches 42.84 dB in 6.52 seconds, surpassing the best baseline result obtained after 1,000 iterations. Within a comparable wall-clock budget, \textit{Ours} continues to improve substantially, reaching 59.12 dB in 22.45 seconds. Exact reconstruction is achieved after 636 iterations, corresponding to 34.17 seconds. Although iterative refinement introduces additional computation per iteration, it yields a substantially better wall-clock fidelity trade-off, reaching higher reconstruction quality much earlier than feed-forward baselines.

\vspace{-10pt}

\subsection{Disentangling Recurrent Refinement and Binarized Supervision}
\label{sec:limitations}
\begin{table*}[t!]
\caption{\textbf{Convergence speed vs.\ fidelity on Kodak-24 (wall-clock).}
Baselines are reported at 1000 optimization iterations, while \textit{Ours} is shown at intermediate checkpoints. PSNR$=\infty$ indicates exact quantized reconstruction.}
\label{tab:supp_timeconsumption}
\centering
\scriptsize
\vspace{-10pt}
\captionsetup[subtable]{
  position=top,
  justification=centering,
  singlelinecheck=false,
  skip=2pt
}

\begin{subtable}[t]{0.49\linewidth}
\centering
\begin{tblr}{
  width = \linewidth,
  colspec = {X[l]ccc},
  row{1} = {font=\bfseries, 
  abovesep = 0pt,
  belowsep = 0pt,},
  rowsep = 2pt,
  colsep = 2pt,
  abovesep = 1pt,
  belowsep = 1pt,
}
\toprule
\textbf{Models} & Iter. & Time (s)$\downarrow$ & PSNR (dB)$\uparrow$ \\
\midrule
$\circ$ Gauss & 1000 & 27.177 & 32.93 \\
$\circ$ SIREN & 1000 & \textbf{21.687} & 33.09 \\
$\circ$ FINER & 1000 & 26.063 & \textbf{40.08} \\
\bottomrule
\end{tblr}
\end{subtable}
\hfill
\begin{subtable}[t]{0.49\linewidth}
\centering
\begin{tblr}{
  width = \linewidth,
  colspec = {X[l]ccc},
  row{1} = {font=\bfseries, 
  abovesep = 0pt,
  belowsep = 0pt,},
  rowsep = 2pt,
  colsep = 2pt,
  abovesep = 1pt,
  belowsep = 1pt,
}
\toprule
\textbf{Models} & Iter. & Time (s)$\downarrow$ & PSNR (dB)$\uparrow$ \\
\midrule
\SetCell[r=3]{l}$\circ$ Ours
& 100  & \textbf{6.522} & \textbf{42.84} \\
& 400  & 22.445 & 59.12 \\
& 636  & 34.167 & $\infty$ \\
\bottomrule
\end{tblr}
\end{subtable}
\vspace{-15pt}
\end{table*}
\begin{wraptable}{r}{0.60\linewidth}
\centering
\vspace{-35pt}
\footnotesize
\setlength{\tabcolsep}{1.9pt}
\renewcommand{\arraystretch}{1.05}
\caption{\textbf{Effect of recurrence and binarized supervision.}
Compared with a feed-forward baseline, recurrent refinement yields a large fidelity gain. Binarized supervision removes the remaining mismatch and enables exact quantized reconstruction.}

\label{tab:supp_threshold_abla}

\resizebox{\linewidth}{!}{%
\begin{tblr}{
  colspec = {l c c c c c},
  row{1}  = {font=\bfseries},
  column{2-6} = {c},
  rowsep  = 2pt,
  colsep  = 4pt,
  row{3}  = {bg=LightBeige,font=\bfseries},
  row{4}  = {bg=OursSoft,font=\bfseries},
}
\toprule
KODAK24 & \#Params. & \#Iteration & PSNR$\uparrow$ & SSIM$\uparrow$ & \#Bit Error$\downarrow$ \\
\midrule
FeedForward~\cite{Liu_2024_CVPR} & 791K   & 1000        & 42.15 & 0.9803 & 305.6K \\
+Rec.                & 593.7K & 1000        & 64.19 & 0.9998 & 49.70K \\
+Rec.+Bin.   & 609K   & $595\pm285$ & $\infty$ & 1.0  & 0 \\
\bottomrule
\end{tblr}}
\vspace{-30pt}
\end{wraptable}
To isolate the architectural benefit of recurrence from the binarized supervision, Table~\ref{tab:supp_threshold_abla} separates the contributions of recurrence and binarized supervision. Relative to the feed-forward baseline, recurrence yields the primary improvement in reconstruction fidelity. Binarized supervision then operates on this high-fidelity recurrent solution, eliminating the remaining mismatch. 
\vspace{-10pt}

\section{Conclusion, Limitations, and Implications}
\label{sec:conclusion}
\vspace{-5pt}
In this work, we offer a spectral perspective on how finite sinusoidal recurrence enriches the spectral support of intermediate INR features.
A harmonic line-spectrum view links sinusoidal features to Fourier structure and provides a spectral interpretation of why recurrent unrolling can expand effective spectral support without increasing the number of parameters. Combined with bipolar code-space supervision, this design yields faster convergence and exact quantized reconstruction on 2D image benchmarks. The same decoder also transfers favorably to super-resolution, NeRF, and SDF tasks. 

\vspace{-15pt}

\subsubsection*{Limitations.}
While our method remains beneficial at moderate-to-large parameter budgets, its gains are not consistently observed across all model sizes, suggesting that its effectiveness may be limited under severe capacity constraints.

\vspace{-15pt}

\subsubsection*{Implications and Future Work.}
Overall, these findings suggest that recurrence is a simple and modular alternative to increasing independently parameterized depth, while understanding and mitigating its model-size dependence remain directions for future work.

\vspace{-8pt}

\subsubsection{Acknowledgments.} This work was partly supported by the National Research Foundation of Korea (NRF) grant funded by the Korea government (MSIT) (RS-2024-00335741), and by the Culture, Sports and Tourism R\&D Program through the Korea Creative Content Agency grant funded by the Ministry of Culture, Sports and Tourism (Project Name: International Collaborative Research and Global Talent Development for the Development of Copyright Management and Protection Technologies for Generative AI; Project Number: RS-2024-00345025), and by the Institute of Information \& Communications Technology Planning \& Evaluation (IITP) grant funded by the Korea government (MSIT) (No. RS-2020-II201336, Artificial Intelligence Graduate School Program (UNIST)).

\newpage

\bibliographystyle{splncs04}
\bibliography{main}
\end{document}